\documentclass[final,5p,times,twocolumn,authoryear]{elsarticle}

\usepackage{amsmath, amssymb, amsfonts}
\usepackage{subfigure}  
\usepackage{float}      
\usepackage{hyperref}   
\usepackage{listings}	

\begin{document}

\begin{frontmatter}

\title{Reinforcement Learning for Control of Valves}

\author{Rajesh Siraskar\corref{cor1}\fnref{fn1}}
\ead{siraskar@coventry.ac.uk; rajeshsiraskar@gmail.com}
\address{Faculty of Engineering, Environment and Computing, Coventry University, UK}
\cortext[cor1]{Corresponding author}
\fntext[fn1]{Author currently works as a Lead Data Scientist at Birlasoft, India.}

\begin{abstract}
This paper is a study of reinforcement learning (RL) as an optimal-control strategy for control of nonlinear valves. It is evaluated against the PID (proportional-integral-derivative) strategy, using a unified framework. RL is an autonomous learning mechanism that learns by interacting with its environment. It is gaining increasing attention in the world of control systems as a means of building optimal-controllers for challenging dynamic and nonlinear processes. Published RL research often uses open-source tools (Python and OpenAI Gym environments). We use MATLAB's recently launched (R2019a) Reinforcement Learning Toolbox\textsuperscript{\texttrademark} to develop the valve controller; trained using the DDPG (Deep Deterministic Policy-Gradient) algorithm and Simulink\textsuperscript{\textregistered} to simulate the nonlinear valve and create the experimental test-bench for evaluation. Simulink allows industrial engineers to quickly adapt and experiment with other systems of their choice. Results indicate that the RL controller is extremely good at tracking the signal with speed and produces a lower error with respect to the reference signal. The PID, however, is better at disturbance rejection and hence provides a longer life for the valves. Successful machine learning involves tuning many hyperparameters requiring significant investment of time and efforts. We introduce ``Graded Learning" as a simplified, application oriented adaptation of the more formal and algorithmic ``Curriculum for Reinforcement Learning''. It is shown via experiments that it helps converge the learning task of complex non-linear real world systems. Finally, experiential learnings gained from this research are corroborated against published research.
\end{abstract}



\end{frontmatter}


\section{Introduction}
Reinforcement learning (RL) is a machine learning technique that mimics learning abilities of humans and animals. RL applications have been used by OpenAI to program robot-hands to manipulate physical objects with unprecedented human-like dexterity \citep{b:OPENAI}, by Stanford's CARMA program for autonomous driving \citep{b:CARMA} and studied for faster de novo molecule design \citep{b:DENOVO}.

This paper studies application of RL as an optimal-control strategy. RL promises better control by interacting directly with the plant and \emph{learning} optimal-control without the need of accurately modeling the plant. 

Valves were selected as the controlled plant as they are ubiquitous in process control and employed in almost every conceivable manufacturing and production industry. Industrial process loops can involve thousands of valves and can be impossible to model accurately. 

PID (proportional-integral-derivative) is the de facto control strategy covering more than 95\% of industrial  controllers \citep{b:DESBOROUGH}. However applying this conventional strategy, can potentially affect quality and efficiency of processes and increase substantial costs.

Connecting a computer to a real physical plant and have the RL agent learn through direct interaction may not always be feasible and a practical approach often adopted, involves simulating the real plant as close as possible and is the approach we use. MATLAB Simulink\textsuperscript{\textregistered} is used to simulate a nonlinear valve, an industrial process, the agent training circuit and finally a unified RL-PID validation circuit. The controller, called an ``agent'' in RL terminology, is trained using MATLAB's recently launched (R2019a) Reinforcement Learning Toolbox\textsuperscript{\texttrademark} using the DDPG (Deep Deterministic Policy-Gradient) algorithm.

\emph{Graded Learning}, a technique discovered accidentally during this research is a simple procedural method to efficiently train a RL agent on complex tasks and in effect is the most simplified form of the more formal method known as ``Curriculum Learning'' \citep{b:NARVEKAR}.\\

\textbf{Summary research contributions of this work}:
\begin{enumerate}
    \item Understanding RL as an optimal-control strategy.
	\item A methodology to assist practising plant engineers apply RL for optimal-control in industries. Design and simulation using MATLAB and Simulink, instead of the more demanding Open Source Python.
    \item \textbf{Graded Learning}: A ``coaching'' method suitable for practicing engineers. It is an application oriented adaptation of the more formal and algorithmic ``Curriculum for Reinforcement Learning''.
    \item Literature research on three studies of RL control for valves.
    \item Experimental comparison of PID and RL strategies in a unified framework.
    \item Stability analysis of the RL controller in time and frequency domains.
    \item Experiential learning corroborated with published research.
\end{enumerate}

Finally, while the valve is the focus of the paper, the methods are adaptable to any industrial system.

\section{Reinforcement Learning Primer}
In this section we take a brief look at conventional optimal-control solving methods, followed by an overview of RL, its connection with optimal-control and finally the DDPG algorithm selected for implementation.

Sutton and Barto's book (\citeyear{b:BARTO}) is the most comprehensive introduction to reinforcement learning and the source for theoretical foundations below.

\subsection{Optimal Control and RL}
Feedback controllers are traditionally designed using two philosophies: adaptive-control and optimal-control. Adaptive controllers are online learners and learn to control unknown systems by measuring real-time data. However, they are not optimized since the design process does not involve minimizing any performance metrics of the plant \citep{b:FRANK}. 

Conventional optimal-control design, on the other hand, is performed off-line by solving Hamilton–Jacobi–Bellman (HJB) equations. According to \citeauthor{b:FRANK} (\citeyear{b:FRANK}) solving HJB equations require complete knowledge of the dynamics of the plant and according to \citeauthor{b:TEDRAKE} (\citeyear{b:TEDRAKE}) this in turn requires an engineered \emph{guess} as a start.

Richard Bellman's extension of the $19^{th}$ century theory laid by Hamilton and Jacobi and Ronald Howard's work in 1960 for solving Markovian Decision Processes (MDPs) all formed the foundations of modern RL. Bellman linked the dynamic system's state with a ``value-function". Dynamic programming, which uses the \emph{Bellman equation} and is a ``backward-in-time'' method, along with \emph{temporal-difference} (TD) methods enabled building of optimal adaptive-controllers for discrete-time systems (i.e. time progression defined as $t, t+1, t+2 ...$) \citep{b:BARTO}.

\subsection{Optimal Control}
The Hamilton-Jacobi-Bellman (HJB) (\ref{eq:HJB}) provides a sufficient condition for optimality \citep{b:TEDRAKE}.

\begin{equation} 
    0 = min _u  \left[ g(x,u) + \frac{\partial J^*}{\partial x}  f (x,u) +  \frac{\partial J^*}{\partial t} \right]
	\label{eq:HJB}
\end{equation}

Controller \emph{policies} (i.e. behavior) are denoted by $\pi$ and optimum policies by $\pi_*$. If a policy $\pi(x,t)$, and a related cost-function $J^\pi(x, t)$, are defined such that $\pi$ minimizes the right-hand-side of the HJB (\ref{eq:HJB}) $\forall x\in\mathbb{R}$ and all $t\in[0,T]$ to zero then:
\begin{equation}
	J^\pi (x,t) = J^* (x,t),\quad \pi(x,t) = \pi^* (x,t)
	\label{eq:HJB_2}
\end{equation}

Equation (\ref{eq:HJB}) assumes that the cost-function is continuously differentiable in $x$ and $t$ and since this is not always the case it does not satisfy all optimal-control problems. In \citep{b:TEDRAKE}, Tedrake shows that solving HJB depends on an engineered \emph{guess}, for example a First-order Regulator is designed with a guessed solution $\pi(x,t) = - sgn(x)$. A Linear Quadratic Regulator is designed similarly. For complex, dynamic mechanical systems such initial solutions are hard to guess unless severely approximated and therefore in situations like these RL shows the relative ease with which real world optimal-control can be \emph{learned}.

\subsection{The RL framework}
\begin{figure}[!ht]
    \centering
    \includegraphics[width=\columnwidth]{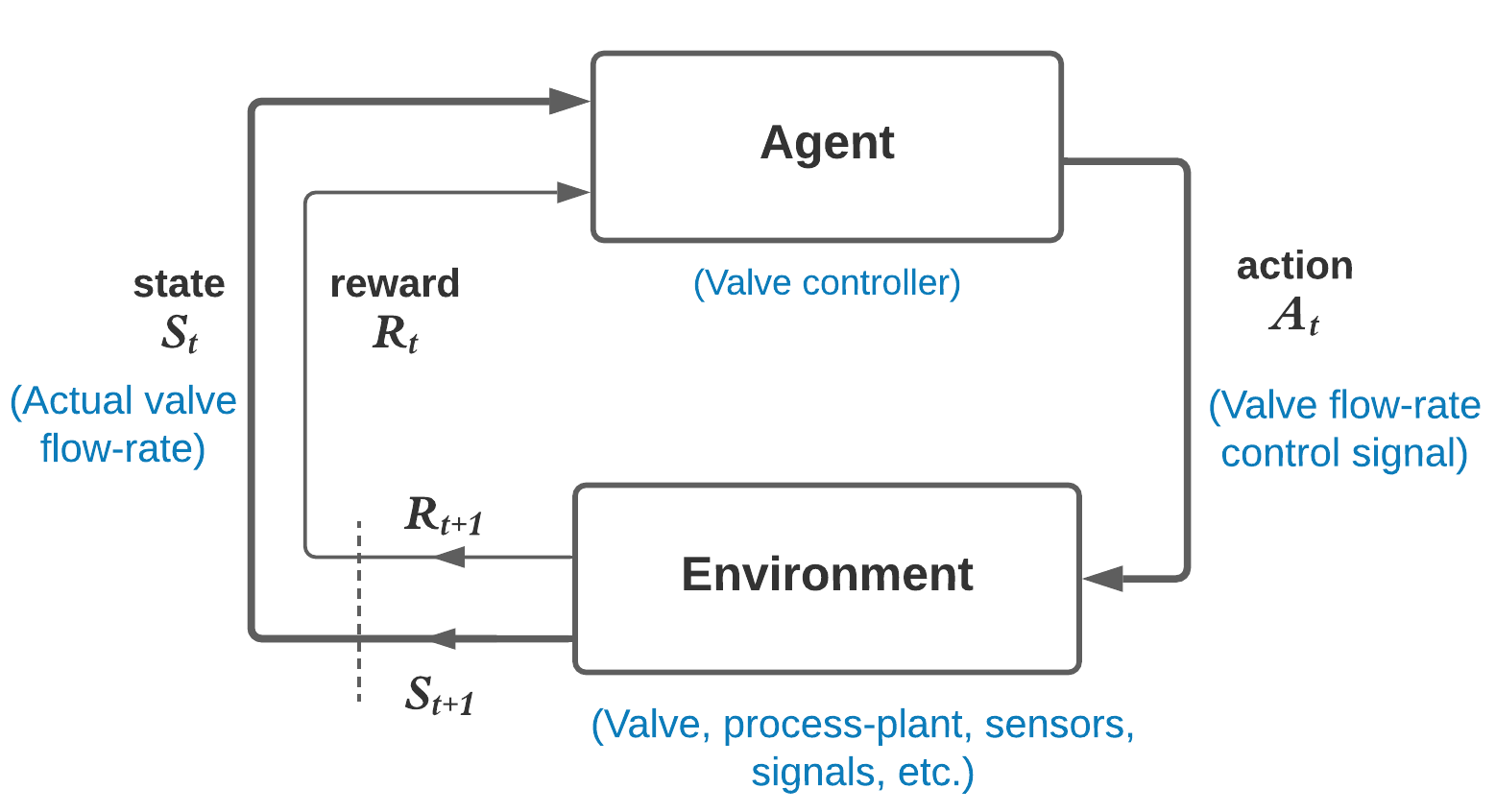}
    \caption{Building blocks of RL and equivalent control-system elements}
    \label{fig:RLFlow}
\end{figure}

The core elements of RL are shown in Fig.\ref{fig:RLFlow} \citep{b_BARTO}, superimposed with equivalent control-system elements. 

The learner and decision-maker is called the \emph{agent}. The agent interacts with its \emph{environment} continually, selecting actions $A_t$, to which the environment responds by presenting a new situation $S_{t+1}$. The environment provides performance feedback via \emph{rewards} (or penalties), $R_{t+1}$. Rewards are scalar values. Over time, the agent attempts to maximize (or minimize) the rewards and this \emph{reinforces} good actions over bad enabling it to learn an optimal \emph{policy}.

In control system terminology the agent is the controller being designed and the environment consists of the system outside the controller i.e. the valve, the industrial process, the reference signal, other sensors, etc. The policy is the optimal-control behavior the designer seeks. RL allows learning this behavior without having to be explicitly programmed or modeling the plant in excruciating detail.

\smallskip
\textbf{Policy}: The decision-making capability of the agent is based on a probability mapping of the best action to take vis-à-vis the state it is in. It is this mapping that forms the policy $\pi_t$, and $\pi_t(a|s)$ is the probability that the action $A_t = a$, if the state $S_t = s$.

\smallskip
\textbf{Returns}: Returns represent long-term rewards, gathered over time.
\begin{equation} 
	G_t  = R_{(t+1)} + R_{(t+2)} + R_{(t+3)} \dots R_T
	\label{eq:returns}
\end{equation}

\smallskip
\textbf{Discounting}: Discounting provides a mechanism to control the impact of selecting an action that is immediate versus one where rewards are received far into the future.
\begin{equation} 
    G_t = R_{(t+1)} + \gamma R_{(t+2)} + \gamma^2 R_{(t+3)} \dots = \sum_{k=0}^{\infty}\gamma^k R_{((t+1)+k)},
    \label{eq:discounting}
\end{equation}
where $\gamma$, the \emph{discount rate}, is a parameter $0 \leq \gamma \leq 1$.

\smallskip
\textbf{Value-functions}: As functions of state-action pairs these provide an estimate of how good it is to perform a given action in a given state. A reward signal provides feedback on how ``good'' the current action is, in an immediate short-term sense. In contrast, a value-function, provides a measure of ``goodness'' in the long-term and is defined in terms of future expected return.

The \emph{value}, denoted $v_\pi(s)$, is the expected return for a state $s$, measured starting in that state $s$ and following the policy $\pi$ thereafter.

\begin{equation}
    \begin{split}
        v_\pi(s) & = \mathbb{E}_\pi [G_t | S_t=s] \\
        & = \mathbb{E}_\pi \left [\sum_{k=0}^{\infty}\gamma^k R_{(t+k+1)} \middle| {S_t=s} \right] \\
        & = \sum_a \pi(a|s) \sum_{s',r} p(s', r|s, a) [r + \gamma v_\pi(s')]
    \end{split}
	\label {eq:bellman}
\end{equation}

Equation (\ref{eq:bellman}) is referred to as the \textbf{Bellman equation} and forms the basis to approximately compute and learn $v_\pi$ and is central to all RL algorithms.

\smallskip
\textbf{Q-function}: By including the action, $q_\pi{(s,a)}$ is defined as the expected return starting from state $s$, taking an action $a$ and thereafter following policy $\pi$.

\smallskip
\textbf{Q-learning}: Q-learning is an off-policy TD control algorithm that allows iteratively learning the Q-value. For each state-action pair; the value $Q(s,a)$ is tracked. When an action $a$ is performed in some state $s$, the two elements of feedback from the environment --- the reward $R$ and the next state $S_{t+1}$ are used in the update shown in (\ref{eq:q-learning}). $\alpha$ is the learning rate.

\begin{multline}
	Q(S_t, A_t) \leftarrow Q(S_t, A_t) + \\
	\alpha [R_{t+1} + \gamma.\max\limits_a Q(S_{t+1}, a) - Q(S_t, A_t)],
    \label{eq:q-learning}
\end{multline}
where $V_t(s)$ is the estimate of $v_{\pi}(s)$ and $Q_t(s,a)$ the estimate of $q_{\pi}(s,a)$.
	
\smallskip
\textbf{Optimal value-function}: There always exists at least one optimal policy that guarantees the highest expected return denoted by $v_*$ and an optimal action-value-function $q_*$.

\smallskip
\textbf{Model-based and model-free RL methods}: Accurate models of the environment allow ``planning'' the next action as well as the reward. An environment's model means having access to a ``table'' of probabilities of being in a state, given an action; and associated rewards. 

RL methods that use environment-models are called \emph{model-based}, as opposed to simpler \emph{model-free} methods. Model-free agents can only learn by trial-and-error \citep{b:BARTO}.

\smallskip
\textbf{Actor-Critic methods}: Actor-critic structure allows a forward-in-time class of RL algorithms that are implemented in real-time. The \emph{actor} component, under a policy, applies an action to the environment and receives a feedback that is evaluated by the \emph{critic}. Learning is a two-step mechanism --- policy-evaluation performed by the critic, followed by policy-improvement, performed by the actor. 
\begin{figure}[!ht]
    \centering
    \includegraphics[width=\columnwidth]{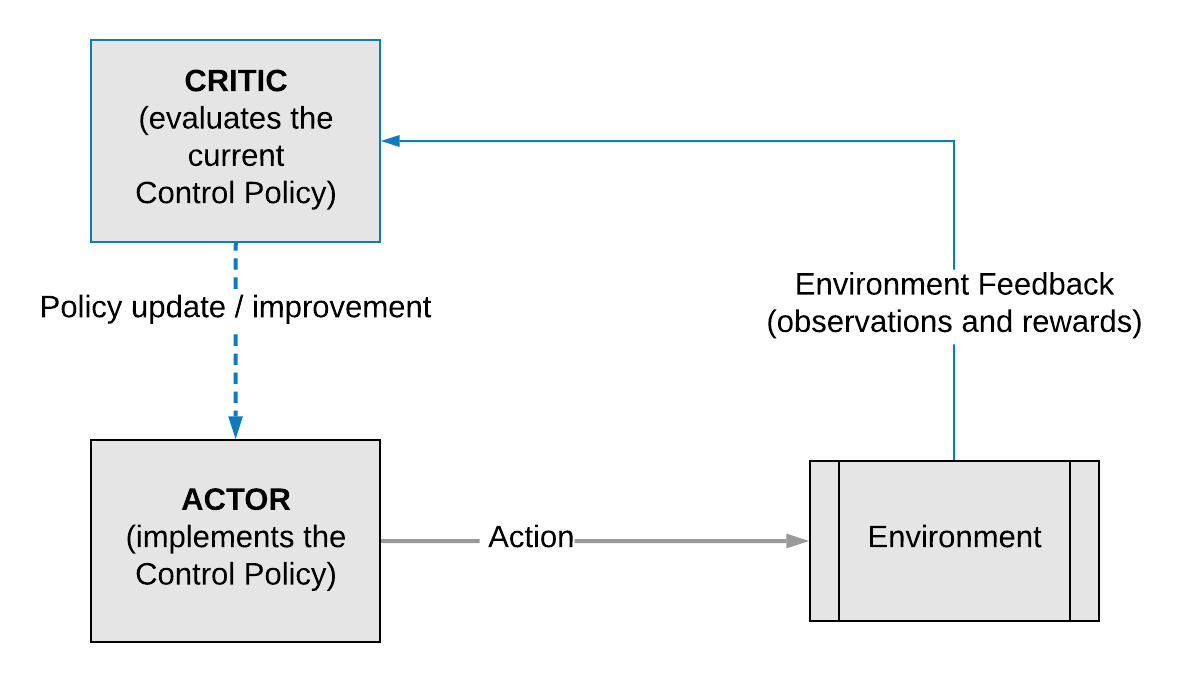}
    \caption{Actor-Critic architecture}
    \label{fig:ACarch}
\end{figure}

\subsection{The DDPG algorithm}
MATLAB's R2019a release provides six RL algorithms. DDPG was the only algorithm suitable for continuous action control \citep{b:MATLAB}.

\citeauthor{b:LCRAP} (\citeyear{b:LCRAP}) introduced DDPG to overcome the shortcomings of the DQN (Deep Q-Network) algorithm which in turn was an extension of the fundamental Q-learning algorithm. 

DDPG is a model-free, policy-gradient and off-policy (by use of a memory replay-buffer of previous \emph{experiences}). As an actor-critic method, it uses two neural-networks. The actor network accepts the current state as the input and outputs a single real value (i.e the valve control signal) representing the action chosen from a continuous action space. The critic network performs the evaluation of the actor’s output (i.e. the action) by estimating the Q-value of the current state given this action. Actor network weights are updated by a \emph{deterministic} policy-gradient algorithm while the critic weights are updated by gradients obtained from the TD error signal. The DDPG algorithm, therefore, simultaneously learns both a Q-function and a policy by interleaving them. 

\smallskip
\textbf{Exploration vs. exploitation}:
For RL, as is in humans, performance improvement is achieved by exploiting the best of past actions. However, to discover these, the agent must first explore untried actions. Balancing this discovery while continuously improving the best action is a common challenge. Various exploration-exploitation strategies have been developed.

DDPG uses the Ornstein-Uhlenbeck process (OUP) to enable exploration. Interestingly, OUP was developed for modeling Brownian particle velocities, with friction, and results in values that are \emph{temporally correlated} \citep{b:OUP}. The simpler additive Gaussian noise model causes abrupt uncorrelated changes from one time-step to another. OUP more closely mimics real life actuators that exhibit inertia \citep{b:LCRAP}.

The exploration policy $\pi'$ is constructed by adding noise to the selected action, sampled from the OUP noise process $\mathcal{N}$.

\begin{equation}
	\pi'(s_t) = \pi(s_t|\theta^\pi) + \mathcal{N}_t
\end{equation}

\section{Control Valves and RL}
Control-valves modify the fluid flow rate, using an actuator mechanism, governed by a control system. Processing plants consist of large networks of control-valves designed to keep a process-variable under control to ensure quality of the end-product \citep{b:ISA}.

\subsection{Nonlinearity in valves}
Control-valves, like most other physical systems, possess nonlinear flow characteristics such as friction and backlash. Friction in-turn has two components --- \emph{stiction}, the static friction, is the inertial force that must be overcome before there is any relative motion between the two surfaces and is the prime cause of dead-band in valves; while dynamic friction is the friction in motion \citep{b:CHOUDHURY_2004_Quantification}, \citep{b:CHOUDHURY_2004_Data_Driven}.

\begin{figure}[!ht]
    \centering \includegraphics[width=0.8\columnwidth]{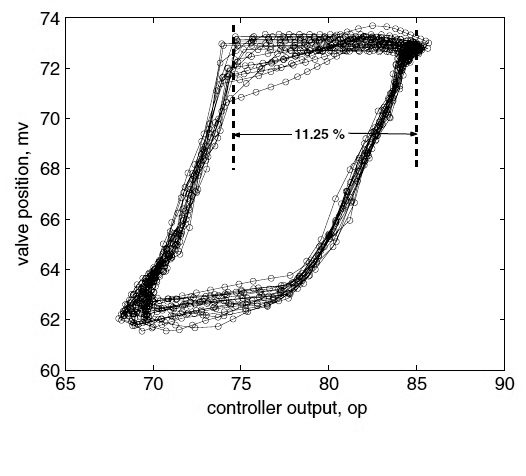}
    \caption{Actual valve movement trajectory \citep{b:CHOUDHURY_2004_Quantification}}
\end{figure}

\begin{figure}[!ht]
    \centering \includegraphics[width=0.7\columnwidth]{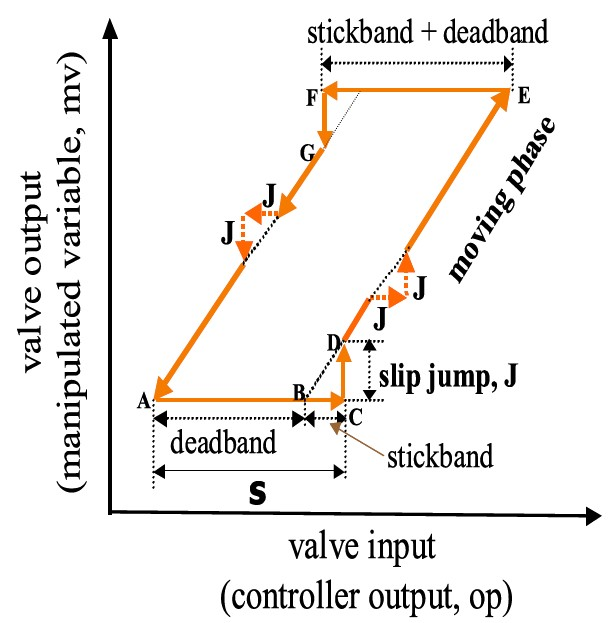}
    \caption{Nonlinear valve operating characteristics, with stiction \citep{b:CHOUDHURY_2004_Quantification}}
\end{figure}

Nonlinearity can cause oscillatory valve outputs that in turn cause oscillations of the process output resulting in defective end-products, inefficient energy consumption and excessive wear of manufacturing systems \citep{b:CHOUDHURY_2004_Quantification}, \citep{b:CHOUDHURY_2005_Modelling}. According to \citeauthor{b:CHOUDHURY_2004_Quantification} (\citeyear{b:CHOUDHURY_2004_Quantification}), 30\% of process-loop oscillation issues are due to control-valves, while \citeauthor{b:DESBOROUGH} (\citeyear{b:DESBOROUGH}) reports that valves are the primary cause of 32\% of surveyed inefficient controllers. Stiction in control-valves has been reported as the prime source of sustained oscillations in industrial control-loops \citep{b:CAPACI}.

\subsection{A mathematical valve model}
RL requires experiences for training. Simulated environments often provide a quick and low-cost environment for training an agent. Since the objective of building a controller is for it to be used in the real-world, one must strive to create as accurate an environment as possible. This appears to contradict the claim made earlier that RL does not require an accurate system model --– however it is assumed here that real physical environment is inaccessible, which on the other hand if accessible or available, could well allow the RL agent (controller) to learn directly from real experiences. 

In this paper we use first-principles to model the valve as outlined in \citep{b:CAPACI}.

\citeauthor{b:HE_2010} (\citeyear{b:HE_2010}) describe the nonlinear memory dynamics of valve by $x_k = N_v (x_{k-1}, u_k)$, at a time-step $k$, where $N_v$ is expressed by relation (\ref{eq:valve_model}). While the controller outputs $u$, the actual position the valve attains is represented by $x$, where $e_k$ represents the valve position error. $f_S$ and $f_D$ are the static (stiction) and dynamic friction parameters, dependent on the valve type, size and application. The ``Experimental Setup'' section describes the Simulink modeling of the valve.

\begin{equation}
    x_k =
    \begin{cases}
        x_{k-1} + [e_k - sign(e_k)f_D], & \text{if}\ |e_k| > f_S \\
        x_{k-1},                       & \text{if}\ |e_k| \leq f_S
    \end{cases}
	\label{eq:valve_model}
\end{equation}
where $e_k = u_k - x_{k-1}$

\subsection{RL for valve control: A literature research}
The field of RL is relatively new and not many studies of its application for control of valves were found \footnote{As of August, 2020, only 18 results showed up for ``reinforcement learning AND valves AND control" on Scopus.}. A study of three publications is presented below with emphasis on overlapping areas of research. 

\subsubsection{Throttle valve control}
Control of a throttle valve is challenging due to the highly dynamic behavior of the spring-damper system and complex nonlinearities \citep{b:BISCHOFF, b:SCHOKNECHT} as well as the multiple-input-multiple-output optimization required \citep{b:HOWELL}.

\citeauthor{b:BISCHOFF} (\citeyear{b:BISCHOFF}) use PILCO (probabilistic inference for learning), a data-efficient \emph{model-based} policy search method. PILCO reduces model bias, a key problem of model-based RL, by learning the probabilistic dynamics of the model and then explicitly incorporating model uncertainty into long-term planning. PILCO is able to learn with only a few trials as against several thousand normally required for trial-and-error based model-free methods \citep{b:PILCO}.

Valve dynamics are modeled using the flap angle, angular velocity and the actuator input. They must be controlled at an extremely high rate of 200 Hz without any overshoot (to avoid engine torque jerks). The controller learns by minimizing the expected sum of cost $c$ over time.

\begin{equation}
    \min\limits_\pi J(\pi), J(\pi) = \sum_{t=0}^{T} \mathbb{E}_{s_t}c(s_t)	
\end{equation}

\begin{figure}[!ht]
    \centering \includegraphics[width=0.6\columnwidth]{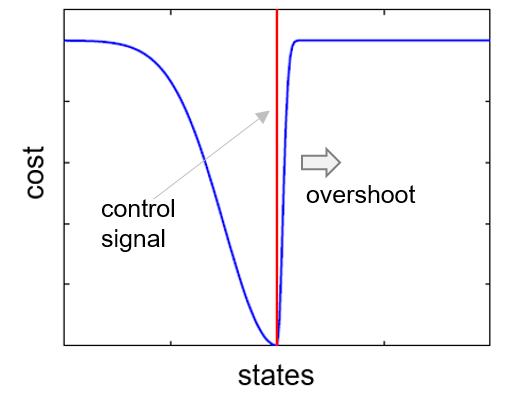}
    \caption{Asymmetric cost-function to avoid overshoots \citep{b:BISCHOFF}}
	\label{fig:Bischoff_CostFunction}
\end{figure}

Fig.\ref{fig:Bischoff_CostFunction} depicts a novel \emph{asymmetric saturating} cost-function applied to achieve the zero-overshoot constraint. A trajectory approaching the goal (red) incurs a rapidly decreasing cost as it nears the goal while overshooting the goal incurs a disproportionately high cost almost immediately \citep{b:BISCHOFF}.

\begin{figure}[!ht]
    \centering
    \subfigure[Control profile]{
        \includegraphics[width=0.8\columnwidth]{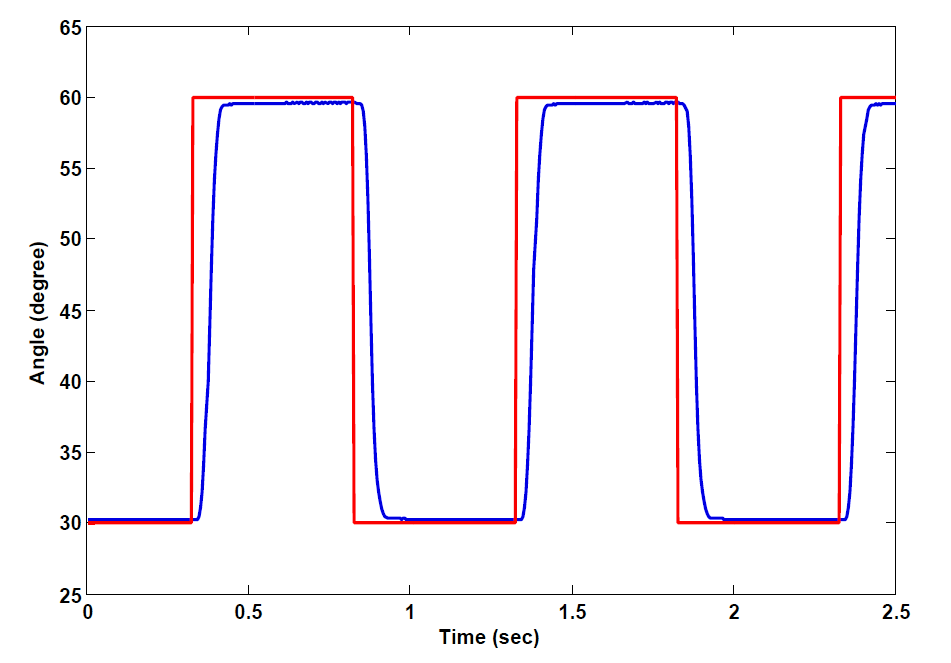}
    }
    \subfigure[Zoomed section shows minor aberrations]{
        \includegraphics[width=0.5\columnwidth]{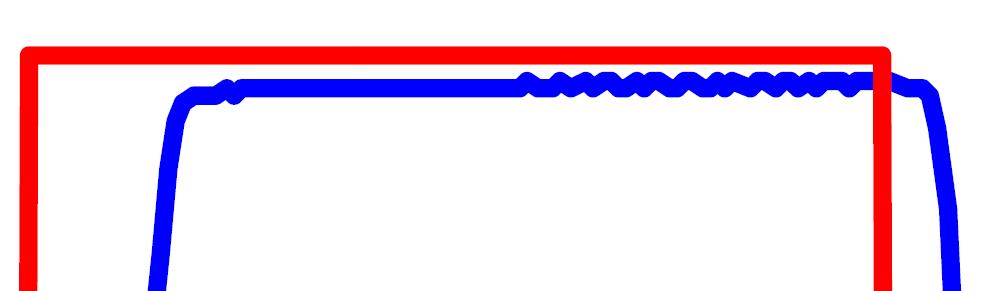}
    }
    \caption{Throttle valve control using PILCO \citep{b:BISCHOFF}}
    \label{fig:Bischoff_Throttle}
\end{figure}

The effectiveness of the cost-function is evident in their results (blue) in Fig.\ref{fig:Bischoff_Throttle}, with no overshoot and only a low-noise behavior of the controlled profile.

\subsubsection{Heating, ventilation and air-conditioning (HVAC) control}
Wang et. al (\citeyear{b:WANG}) use a model-free, proximal actor-critic based RL algorithm to control the nonlinear dynamics of HVAC systems where the hot-water flow is governed by a $3^{rd}$ power equation (\ref{eq:Wang}).

\begin{multline}
    f{_w}(t) = 0.008 + 0.00703(-41.29 + 0.309{ }u - \\ 
    0.368 \times 10^{-4} u^2 + 9.56 \times 10^{-8}u^3)
    \label{eq:Wang}
\end{multline}
	
RL is compared to Proportional-Integral (PI) and Linear Quadratic Regulator (LQR) control strategies. 150 time-steps are used to allow sufficient time for RL controller to learn tracking the set-point. Disturbances are simulated using random-walk algorithm. Actor network configuration is \texttt{[50, 50]} and the critic is a single layer of 50 units. One interesting aspect of the network architecture they employ is the use of GRU (Gated Recurrent Unit) to overcome the problem of vanishing/exploding gradients.

\begin{figure}[!ht]
    \centering \includegraphics[width=\columnwidth]{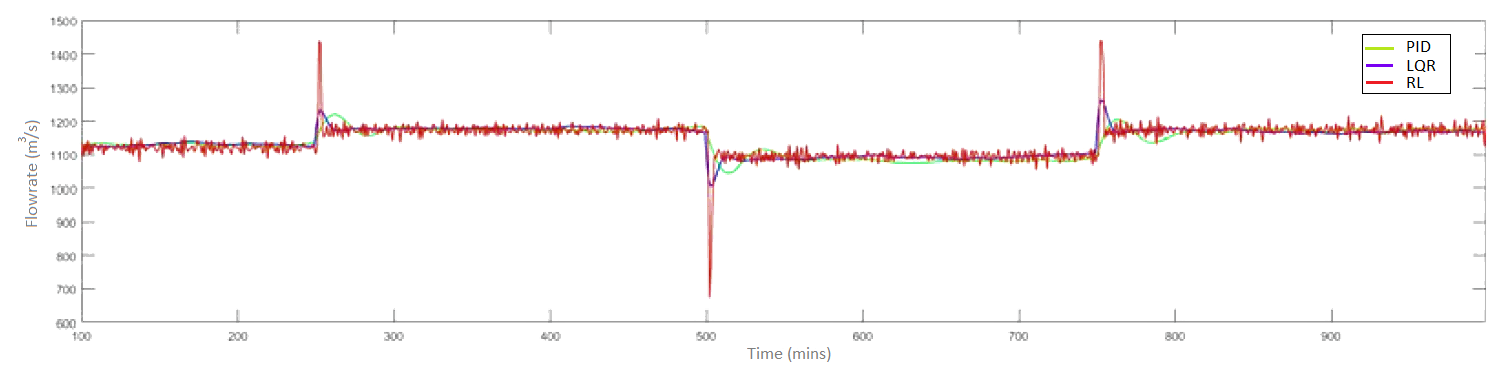}
    \caption{HVAC control \citep{b:WANG}}
    \label{fig:HVAC_Wang}
\end{figure}

Fig.\ref{fig:HVAC_Wang} shows that the RL controller responds much faster than the LQR and PI controllers and tracks the reference signal better with lower integral-absolute and integral-square errors. However the RL shows a very high-variance noisy response against the smooth trajectories of PI and LQR controllers. Significant overshoots are also seen in the RL response.

\subsubsection{Sterilization of canned food}
Thermal processing used for sterilization of canned food results in deterioration of the organoleptic properties of the food. Controlling the thermal process is therefore important. \citeauthor{b:SYAFIIE} (\citeyear{b:SYAFIIE}) apply Q-learning to learn the optimal temperature profile that can be applied during the two stages of the thermal process --- manipulation of the saturated-steam valve to cause heating and then cooling by opening the water valve. 

\begin{figure}[!ht]
    \centering
    \subfigure[Control profile]{
        \includegraphics[width=0.8\columnwidth]{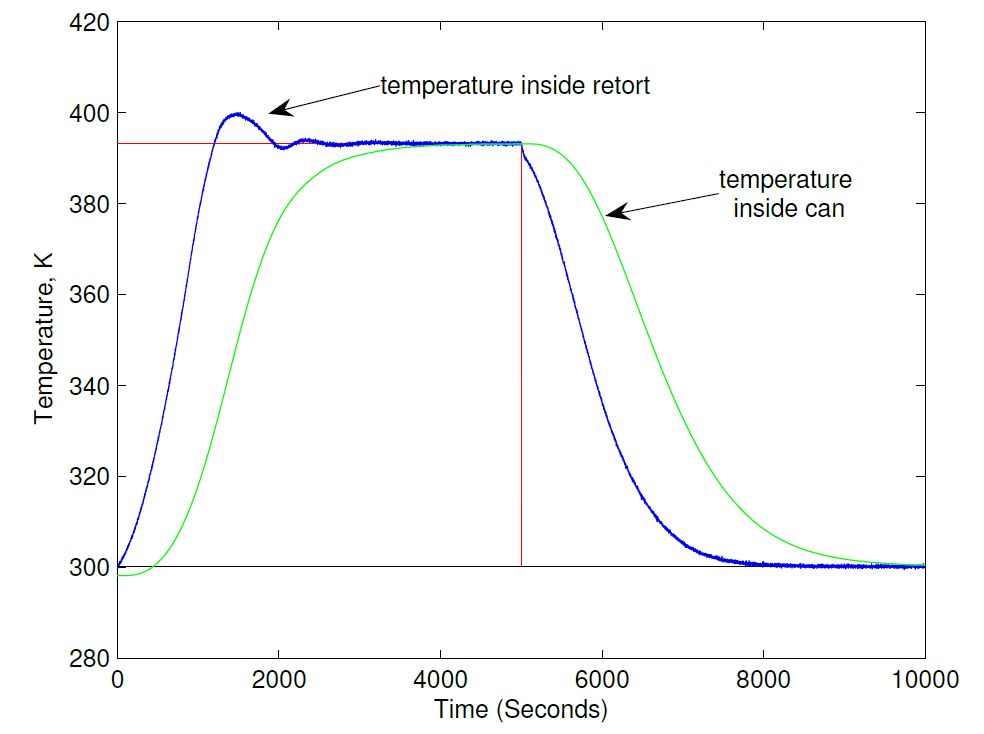}
    }
    \subfigure[Zoomed section shows aberrations]{
        \includegraphics[width=0.5\columnwidth]{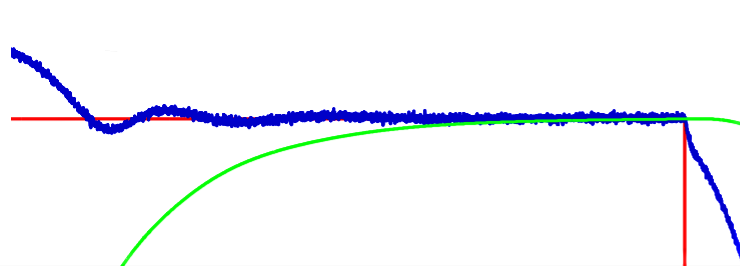}
    }
    \caption{Thermal process control using Q-learning \citep{b:SYAFIIE}}
    \label{fig:Syafiie_Temperature}
\end{figure}

A simple \emph{scalar} reward is used \texttt{[+1.0, 0.0, -2.0]}, therefore penalizing an action deviating from the desired start twice as more as rewarding it. The paper does not evaluate continuous rewards. Fig.\ref{fig:Syafiie_Temperature} shows the controlled temperature profile.

\textbf{Overall observations on the three researched papers:} 
\begin{enumerate}
	\item Disturbances in the RL controlled signal are evident in all three implementations;  \citep{b:BISCHOFF}, \citep{b:WANG} and \citep{b:SYAFIIE}.
	\item Use of stochasticity mechanisms other than OUP to enable exploration of action space; \citep{b:BISCHOFF} and \citep{b:WANG}.
	\item Use of a novel objective function in \citep{b:BISCHOFF}.
	\item \emph{None} of these evaluated the stability of the RL controller design --- an important consideration for an emerging breed of controllers.
	\item MATLAB was not used for RL design\footnote{MATLAB launched the Reinforcement Learning Toolbox\textsuperscript{\texttrademark} in Mar. 2019 \citep{b:MATLAB}}.
	\item Only \citeauthor{b:WANG} (\citeyear{b:WANG}) compared the RL against the traditional PID.
\end{enumerate}

\section{Experimental Setup}
\begin{figure}[!ht]
    \centering 
	\includegraphics[width=0.8\columnwidth]{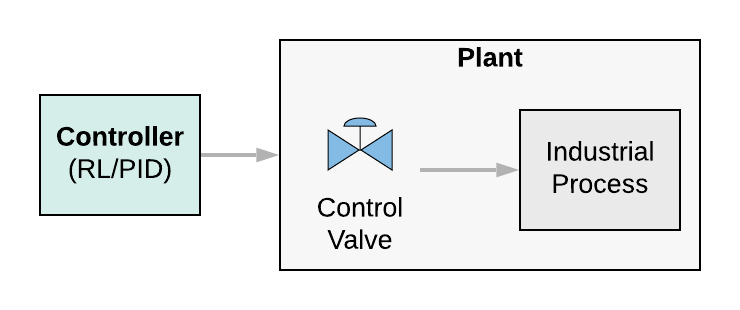}
    \caption{Basic block components}
	\label {fig:basic_setup}
\end{figure}

This section describes the creation of the experimental setup, using MATLAB and Simulink, for design and evaluation of the RL and PID controllers. Fig.\ref{fig:basic_setup} shows the core components. 

Our setup used elements from the excellent \citeyear{b:CAPACI} paper, \emph{``An augmented PID control structure to compensate valve stiction"} by \citeauthor{b:CAPACI}. 

\begin{figure}[!ht]
    \centering
    \subfigure[Two-move compensator]{
        \includegraphics[width=\columnwidth]{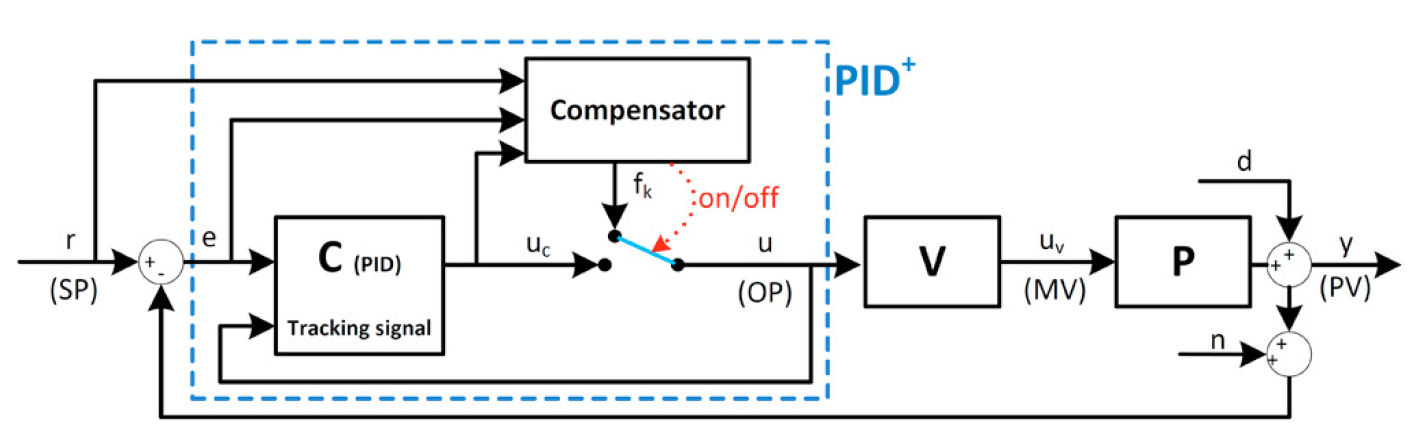}
        \label{fig:Capaci_PID}
    }
	\subfigure[Compensator results on a constant reference signal]{
        \includegraphics[width=\columnwidth]{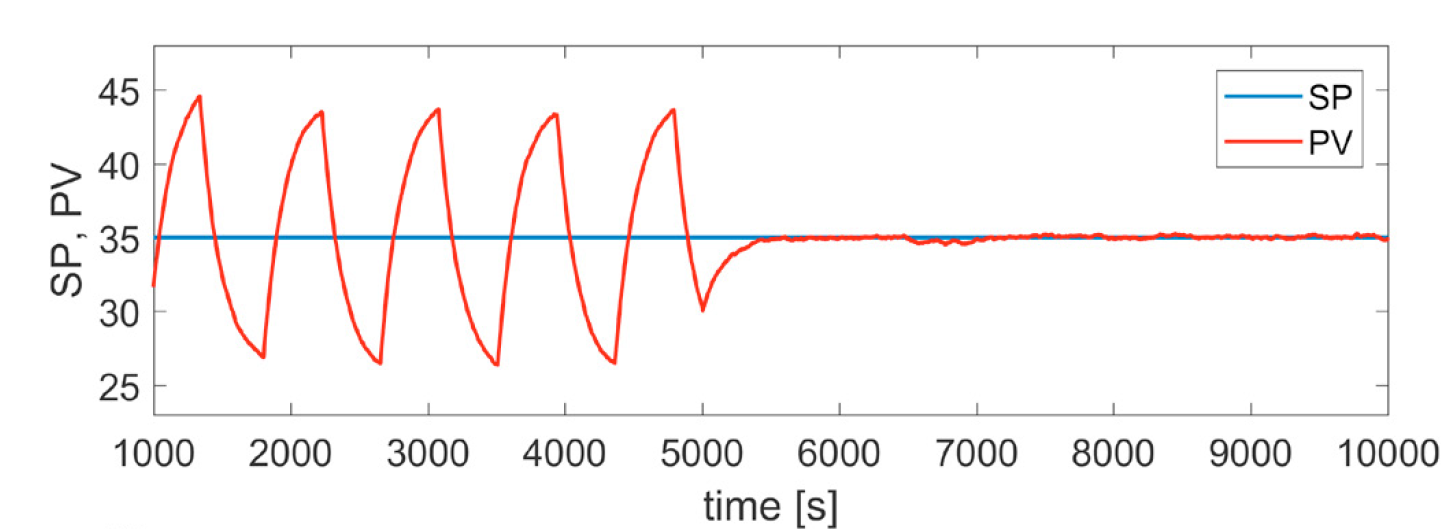}
        \label{fig:Capaci_results}
    }
	\subfigure[Compensator results on a process with loop perturbations]{
        \includegraphics[width=\columnwidth]{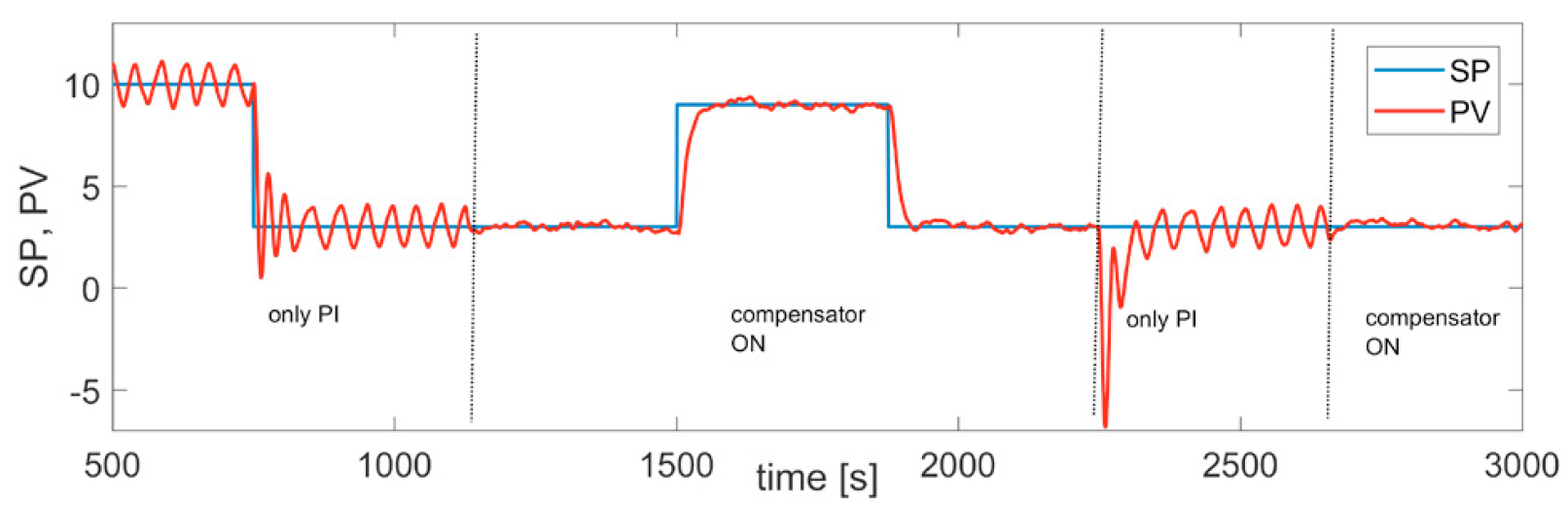}
        \label{fig:Capaci_Pertubations}
    }
	\subfigure[Regenerated ``benchmark waveform"]{
        \includegraphics[width=0.8\columnwidth]{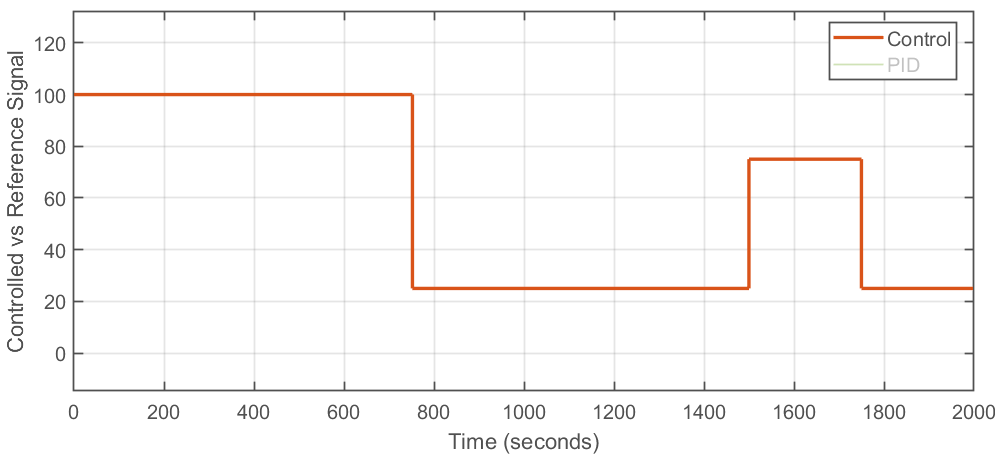}
        \label{fig:Capaci_waveform}
    }
    \caption{``PID compensator'' by \citeauthor{b:CAPACI} (\citeyear{b:CAPACI})}
    \label{fig:Capaci}
\end{figure}

Traditional PID controllers tuned solely on process dynamics, cause sustained oscillations attributed to the integral component that causes excessive variation of the control action to overcome static friction \citep{b:CAPACI}. As a solution to this \citeauthor{b:CAPACI} (\citeyear{b:CAPACI}) presented a novel PID based controller, Fig.\ref{fig:Capaci_PID}, where stiction is overcome by employing a \emph{two-move control sequence} (\ref{eq:Capaci_equations}) as the valve input. 

\begin{equation}
    \begin{split}
    u_k &=
    \begin{cases}
        u_{k-1} + a\hat f_S, & \text{if } u_{k-1} \geq \hat x_{ss} \\
		u_{k-1} - a\hat f_S, & \text{if } u_{k-1} < \hat x_{ss} \\
    \end{cases}\\
	u_{k+1} &=
    \begin{cases}
	    \hat x_{ss} - \hat f_D, & \text{if } u_{k-1} \geq \hat x_{ss} \\
	    \hat x_{ss} + \hat f_D, & \text{if } u_{k-1} < \hat x_{ss} \\
    \end{cases}\\
	u_{k+j} &= u_{k+1} (=u_{ss}), \text{ if } j > 1
	\end{split}
	\label{eq:Capaci_equations}
\end{equation}
where $\hat f_S$ and $\hat f_D$ are estimates of stiction and dynamic friction and $\hat x_{ss}$ is the estimate of steady-state position of the valve. Equation (\ref{eq:Capaci_equations}) is highly dependent on the accurate estimations of friction parameters.\\

\textbf{The setup components}:
\begin{enumerate}
    \item A PID (with filter) controller tuned using MATLAB's auto-tuning feature.
	\item A training setup for the RL agent using the DDPG algorithm.
	\item A unified framework for experimentation and evaluation of controllers \\
	
	Items below were based on \citep{b:CAPACI}:
	\item Nonlinear valve model (\ref{eq:Capaci_equations}) including the valve friction values $f_S$ and $f_D$.
	\item Two industrial processes controlled by the valve:
	\begin{enumerate}
		\item Normal process (\ref{eq:FOPTD}).
		\item Process with loop perturbations (\ref{eq:perturbation}).
	\end{enumerate}	
	\item A ``benchmark waveform'' profile with noise parameters (Fig.\ref{fig:Capaci_waveform}).
\end{enumerate}

\subsection{Modeling the valve}
Simscape Fluids\textsuperscript{\texttrademark} provides simulations for several valve types and is the simplest and quickest option. \citep{b:POPINCHALK} is a MathWorks article to enhance these into more realistic models using an understanding of system dynamics.

\begin{figure}[!ht]
    \centering 
	\includegraphics[width=\columnwidth]{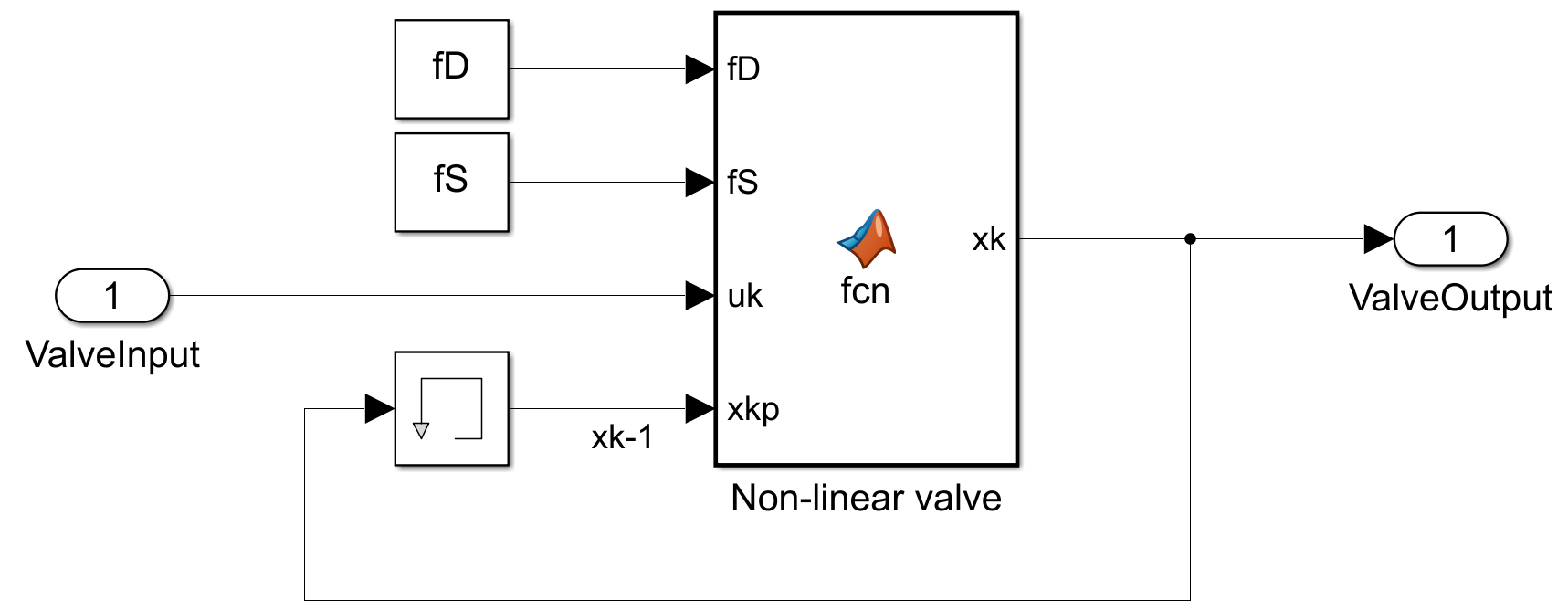}
    \caption{Simulink valve model}
	\label {fig:valve_model}
\end{figure}

We use first-principles and mathematically model the nonlinear valve. Algebraically rearranging equations shown in (\ref{eq:Capaci_equations}) produces (\ref{eq:valve_model_rearrange}); these equations are then implemented in Simulink using code in a ``user-defined-function'' (Listing.\ref{code:valve_code}) and a ``memory'' block shown in Fig.\ref{fig:valve_model} with $f_S=8.40$ and $f_D=3.524$.

\begin{equation}
    x_k =
    \begin{cases}
        u_k - f_D, & \text{if}\ u_k  - x_{k-1} > f_S \\
        u_k + f_D, & \text{if}\ u_k  - x_{k-1} < -f_S \\
        x_{k-1},   & \text{if}\ |u_k  - x_{k-1}| \leq f_S
    \end{cases}
	\label{eq:valve_model_rearrange}
\end{equation}

\begin{lstlisting}[caption=MATLAB script for modelling a non-linear valve, label=code:valve_code, fontadjust, basicstyle=\small]
function xk = fcn(fD, fS, uk, xkp)
    t_xk = 0.0;
    if ((uk-xkp) > fS)
        t_xk = uk - fD;
    elseif ((uk-xkp) < -1*fS)
        t_xk = uk + fD;
    elseif (abs(uk-xkp) < fS)
        t_xk = xkp;
    end
    
    % Return the valve output
    xk = t_xk;
\end{lstlisting}

\subsection{Modeling the ``industrial'' process}
\begin{figure}[!ht]
    \centering \includegraphics[width=\columnwidth]{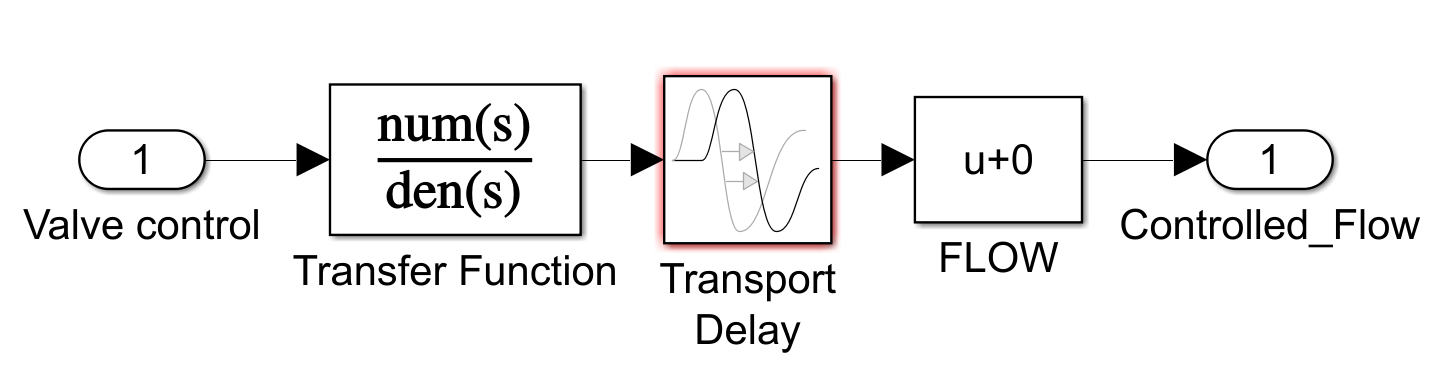}
    \caption{FOPTD process model}
	\label{fig:FOPTD_model}
\end{figure}

The benchmark ``industrial process'' is modeled as a first-order plus time-delay (FOPTD) process (\ref{eq:FOPTD}),  using transfer-function and time-delay blocks as shown in Fig.\ref{fig:FOPTD_model}.
\begin{equation} 
	G(s) = \frac {k} {(1+Ts)} e^{-Ls},
	\label{eq:FOPTD}
\end{equation}
where $k=3.8163$, $T=156.46$ and $L=-2.5$.

\subsection{PID controller setup}

\begin{figure}[!ht]
    \centering \includegraphics[width=\columnwidth]{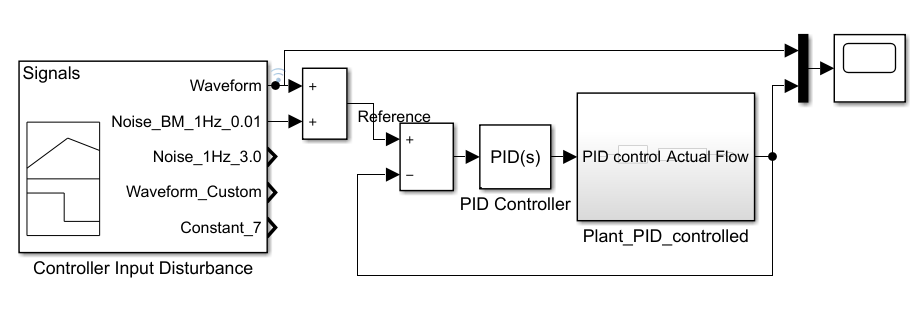}
    \caption{PID control setup}
\end{figure}

A PID controlled output is a function of the feedback error, represented in time-domain as:

\begin{equation}
    u(t) = K_pe + K_i\int e(t).dt + K_d \frac {de}{dt}
	\label{eq:pid_ideal}
\end{equation}
where $u$ is the desired control signal and $e(t) = r(t) - y(t)$ is the tracking error, between the desired output $r$ and the actual output $y$. This error signal is fed to the PID controller, and the controller computes both the derivative and the integral of this error signal with respect to time providing a set-point tracking effect, this works continuously in a closed loop, until the controller is in effect.

The ideal theoretical PID form exhibits a drawback for high frequency signals --- the derivative action results in very high gain. A high frequency measurement noise will therefore generate large variations in the control signal. Practical implementations reduce this effect by replacing the $K_d$ term by a first-order filter (where $K_d.de/dt$ is represented as $K_d.s$ in Laplace form) by (\ref{eq:pid_practical}) \citep{b:MURRAY}.

\begin{equation}
    K_p + K_i \frac {1}{s} + K_d \frac{N}{(1 + N \frac{1}{s})}
	\label{eq:pid_practical}
\end{equation}

The filter coefficient $N$ determines the pole location of the filter that helps attenuate the high gain on high-frequency noise. A $N$ between $2$ and $20$ is recommended. A high value ($N>100$) results in (\ref{eq:pid_practical}) approaching the ideal form (\ref{eq:pid_ideal}) \citep{b:MURRAY}. 

The PID was tuned using MATLAB auto-tuning feature and the coefficients obtained were $K_p=0.3631$, $K_i=0.0045$, $K_d=-1.72$ and $N=0.0114$. The low $N$ acts to suppress the derivative term.

\subsection{RL controller setup}
\begin{figure}[!ht]
    \centering \includegraphics[width=\columnwidth]{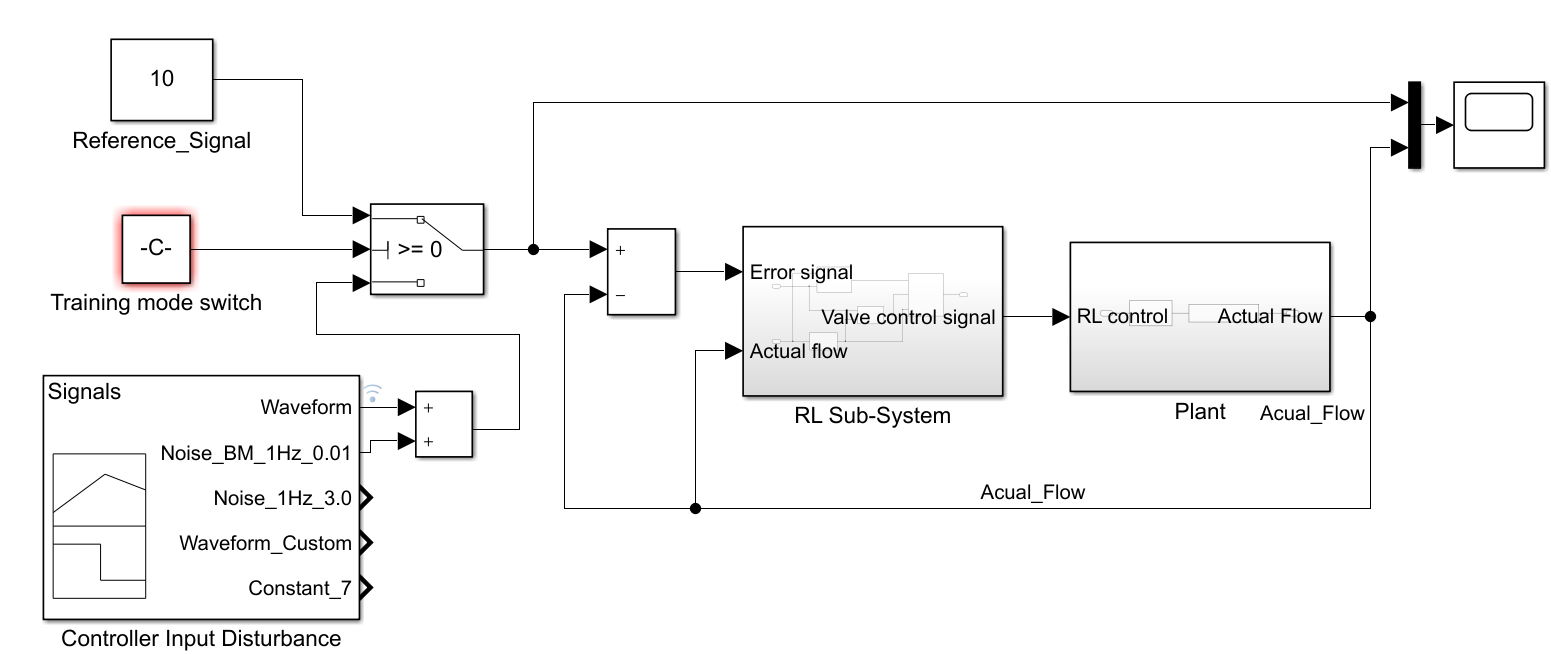}
    \caption{RL agent training setup}
	\label{fig:RL_Training}
\end{figure}

Fig.\ref{fig:RL_Training} shows the Simulink setup, for training \emph{and} evaluation of the RL controller.

Training an agent involves significant hyperparameter tuning and a switch allowed for quick experiments with numerous signals fed via a ``signal-builder'' block.

\subsubsection{RL controller design}
\begin{figure}[!ht]
    \centering \includegraphics[width=\columnwidth]{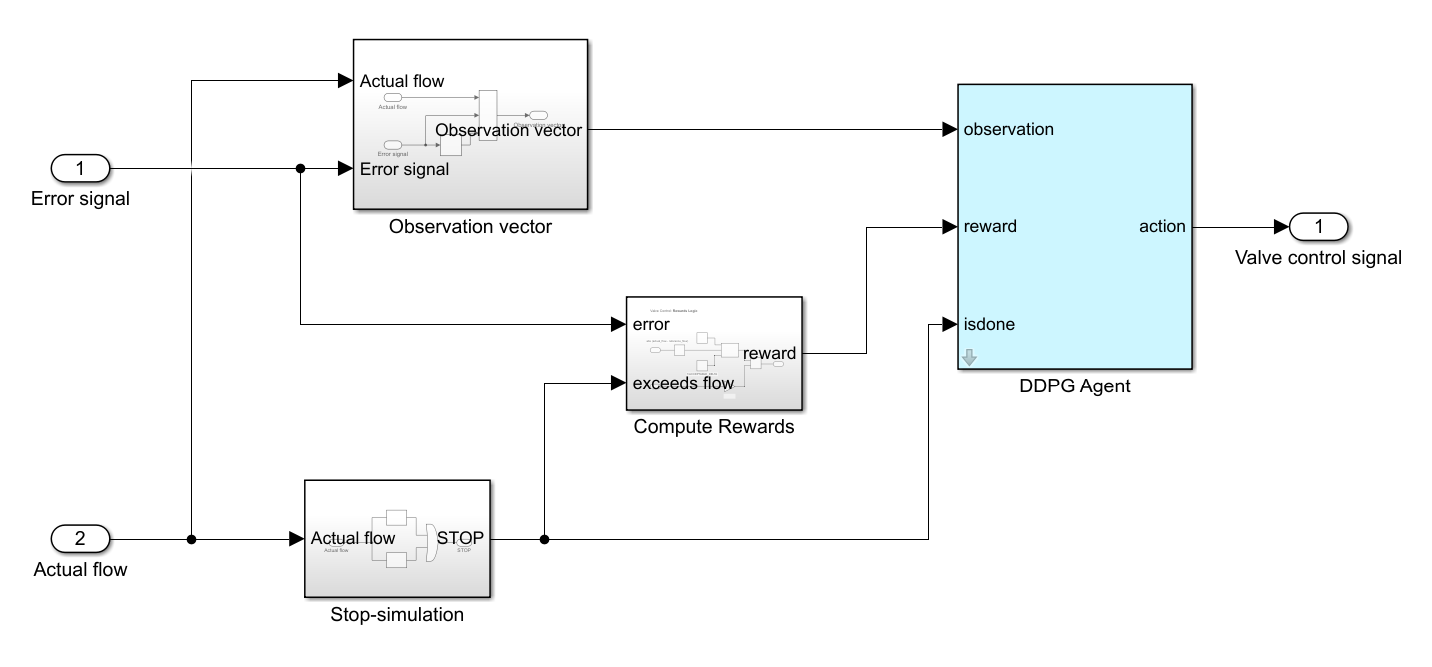}
    \caption{RL DDPG agent details}
	\label{fig:RL_Agent_Block} 
\end{figure}

Fig.\ref{fig:RL_Agent_Block} shows the DDPG Agent block with feedback from the environment channelized via the Observations vector. It also shows the block that computes Rewards and the Stop-simulation block that controls the termination of an episode. 

\subsubsection{Environment design}
Several design factors need consideration, when building the environment, for efficiently training an agent to follow trajectories of a control signal. They can broadly be classified into agent related and environment related.

Agent related factors are composition of the observations vector and the reward strategy. Environment related factors cover the training strategy, training signals, initial conditions of the environment and criteria to terminate an episode.

\subsubsection{Training strategy}
One could train the RL agent to follow the exact benchmark trajectory (Fig.\ref{fig:Capaci_waveform}), however this is a very constrained strategy. Instead, the agent was trained to follow random levels of straight-line (constant) signals. The agent was additionally challenged to learn to start at a randomly initialized flow value. Together this forms an effective and generalized training strategy to teach the agent to follow \emph{any} control signal trajectory composed of straight lines. The RL ToolBox allows overriding the default ``reset function'' that assists in implementing the above strategy.

\begin{verbatim} 
 env.ResetFcn = @(in)localResetFcn(in, 
                VALVE_SIMULATION_MODEL); 
\end{verbatim}

\subsubsection{Observation vector}
\begin{figure}[!ht]
    \centering \includegraphics[width=\columnwidth]{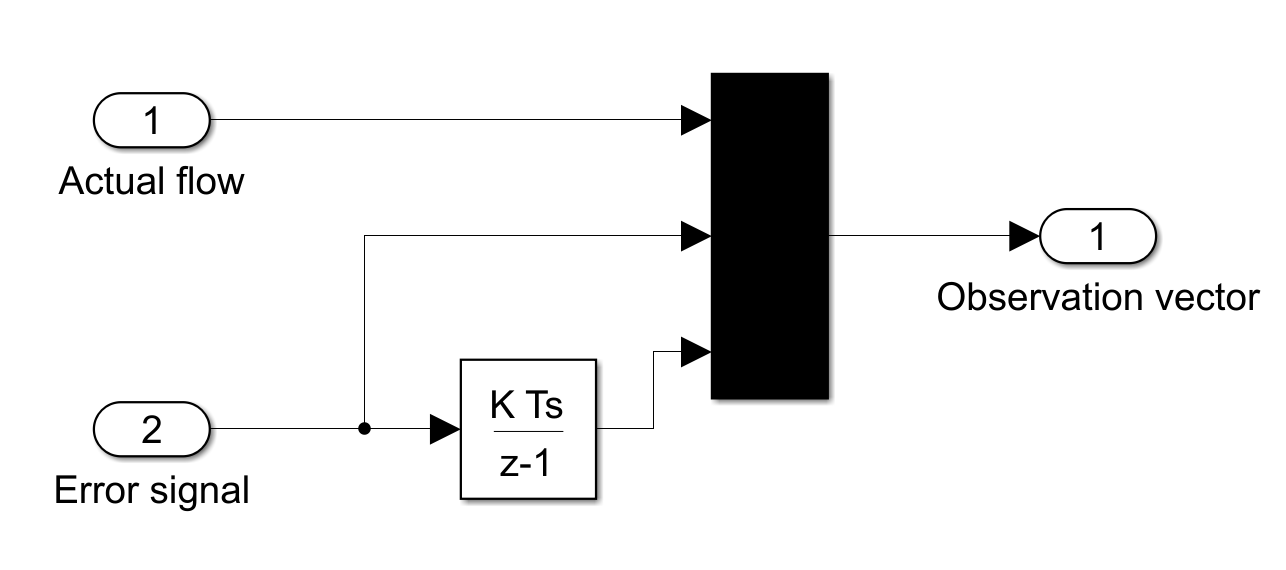}
    \caption{RL observations vector}
	\label{fig:observation_vector} 
\end{figure}

The observation vector, modeled as shown in Fig.\ref{fig:observation_vector}, is composed of: $[y; e; \int e.dt]^T$, where $y$ is the actual flow achieved, $e$ the error with respect to reference $r$, and finally the integral of the error.

\begin{figure}[!ht]
    \centering \includegraphics[width=0.5\columnwidth]{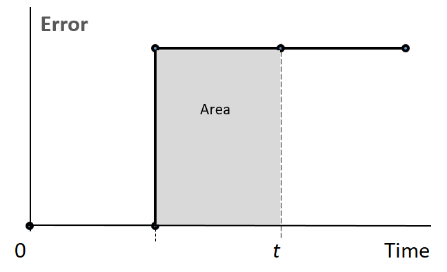}
    \caption{Error integral}
	\label{fig:error_integral}
\end{figure}

\smallskip
\textbf{Integral of error}: The instantaneous error has no memory. The integral of error, which is the area under the curve as time progresses, provides a mechanism to compute the total error gathered over time and drive the agent to lower this (Fig.\ref{fig:error_integral}). This is an important observation input often used in training of RL controllers. 

\subsubsection{Rewards strategy}
Rewards can be assigned via discrete, continuous or hybrid functions. Equation (\ref{eq:scalar_reward}) is a simple discrete form.

\begin{equation}
    Reward = 
    \begin{cases}
        10, & \text{if}\ |e| < \Delta\\
        -1, & \text{if}\ |e| \geq \Delta\\
        -100, & \text{if}\ (y \leq 0,  y > Max\_Flow),
    \end{cases}
	\label {eq:scalar_reward}
\end{equation}
where $\Delta$ is some allowable error margin.

Equation (\ref{eq:cont_reward}) shows a reward that varies continuously as a function of error $e$. $\lambda$ is a small constant that avoids division-by-zero error.
\begin{equation}
    Reward = 
    \begin{cases}
        -100, & \text{if}\ (y \leq 0,  y > Max\_Flow)\\
        \frac{1} {(e+\lambda)} & \text{otherwise}
    \end{cases}
	\label {eq:cont_reward}
\end{equation}

\begin{figure}[!ht]
    \centering \includegraphics[width=\columnwidth]{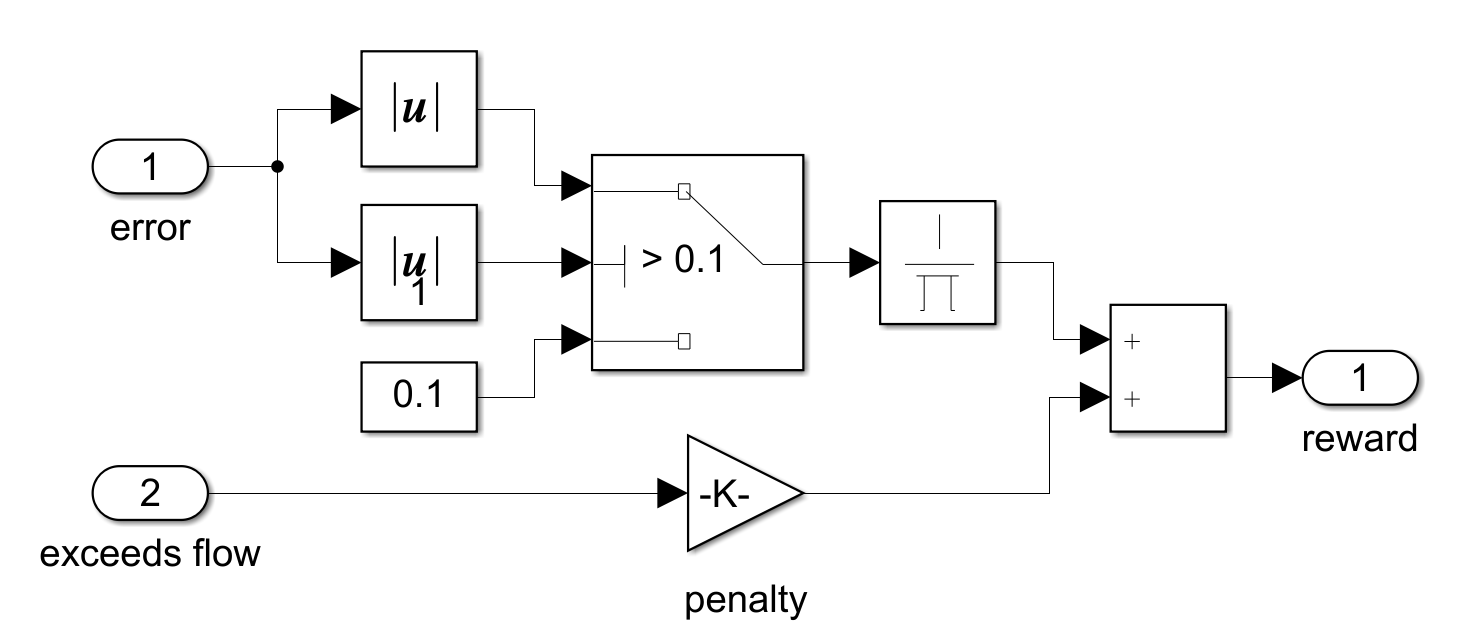}
    \caption{RL rewards computation block}
	\label{fig:RewardsComputation} 
\end{figure}

Fig.\ref{fig:RewardsComputation} shows the final implementation as a hybrid form. The reciprocal of the absolute error allows the controller to learn to drive the error lower and lower. The discrete part of the reward is the ``penalty'' block that assigns a set penalty for exceeding the flow limits.

\subsubsection{Actor and Critic networks}
\begin{figure}[!ht]
    \centering
    \subfigure[Policy (actor) network]{
        \includegraphics[width=0.6\columnwidth]{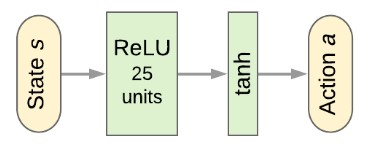}
    }
    \subfigure[Critic (action-value) network]{
        \includegraphics[width=0.8\columnwidth]{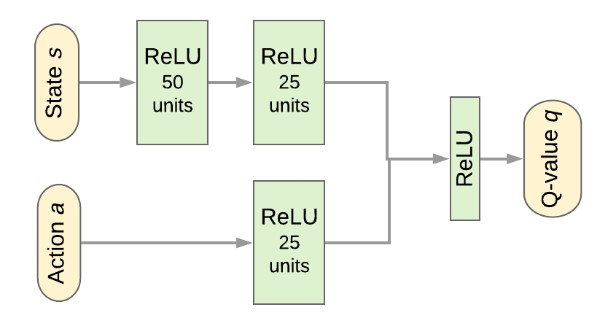}
    }
    \caption{DDPG network architectures}
    \label{fig:DDPG_architecture}
\end{figure}

The actor-critic DDPG components were implemented as shown in Fig.\ref{fig:DDPG_architecture}. The networks have fully-connected layers, initialized with small random weights before beginning the training. 

The actor network output is normalized to be between \texttt{[-1, 1]} using a \texttt{tanh} layer. This allows better learning and convergence for continuous action spaces.

\subsubsection{Ornstein-Uhlenbeck (OU) action noise parameters}
\citep{b:MATLAB_DDPG} provides guidelines for computing the DDPG exploration parameters --- noise model variance and its decay rate via (\ref{eq:oup_var}).

\begin{equation}
	Variance.\sqrt{T_s} = (1\%\text{ to }10\%) \text{ of } ActionRange
	\label{eq:oup_var}
\end{equation}
where $Ts$ is the sampling time\newline

On deciding a half-life of the variance factor, in time-steps, the decay rate is then computed using (\ref{eq:oup_halflife})

\begin{equation}
    HalfLife = \frac {\log (\frac{1}{2})} {\log(1 - VarianceDecayRate)}
	\label{eq:oup_halflife}
\end{equation}

\subsubsection{Final DDPG hyperparameters}
Table \ref{table:DDPG_hyperparameters} summarizes the final set of DDPG hyperparameters.
	
\begin{table}[H]
	\caption{DDPG hyperparameter settings}
    \renewcommand{\arraystretch}{1.5}
	\begin{center}
	\begin{tabular}{ |l|l| } 
		\hline
		Hyperparameter & Setting \\
		\hline
		Critic learning rate	& 1$e^{-03}$ \\
		Actor learning rate		& 1$e^{-04}$ \\
		Critic hidden layer-1	& 50 fully-connected \\
		Critic hidden layer-2	& 25 fully-connected \\
		Action-path neurons		& 25 fully-connected \\
		Action-path bound		& tanh layer\\
		Gamma					& 0.9 \\
		Batch size				& 64 \\
		OUP Variance 			& 1.5 \\
		OUP Variance Decay Rate	& 1$e^{-05}$ \\
		\hline
	\end{tabular}
	\end{center}
	\label{table:DDPG_hyperparameters}
\end{table}

\subsection{Setup for comparative study}
\begin{figure*}
	\centering 
	\includegraphics[width=\textwidth]{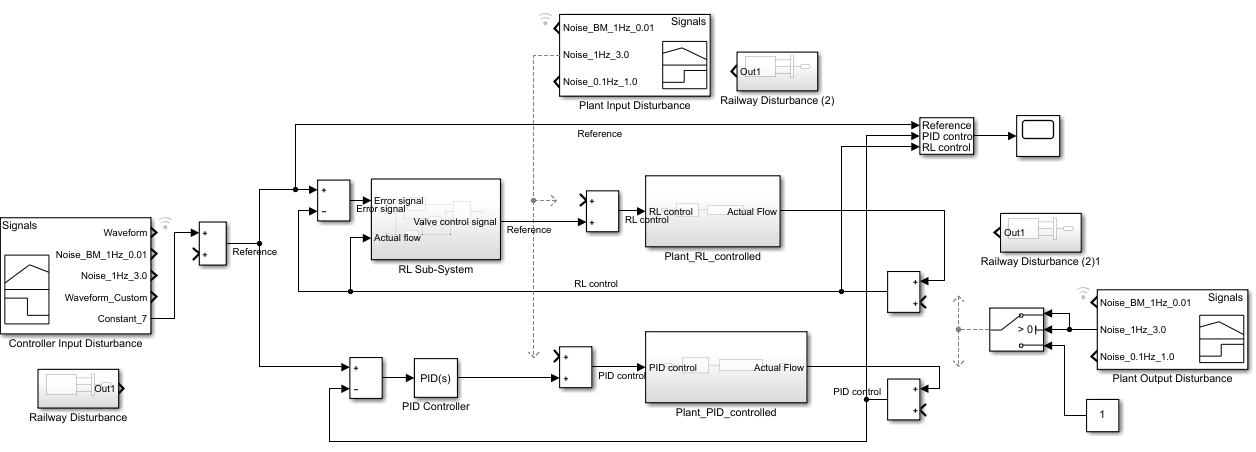} 
	\caption{Unified setup for a comparative evaluation of RL and PID control strategies}
	\label{fig:EvaluationSetup}
\end{figure*}

An environment that combines the PID and RL strategies for a comparative evaluation is shown in Fig.\ref{fig:EvaluationSetup}. It allows experimenting with various reference signals, studying the effects of noise added at three disturbance points i.e. input of the controller, output of the controller (i.e. the input of the plant) and finally output of the plant.

It provides a convenient platform to perform additional experiments using elements such as set-point filters, output smoothing filters, etc.

\section{Graded Learning}
Before presenting the results of the experiments we elaborate on ``Graded Learning'', a progressive \emph{coaching method}. This simple, \emph{intuition} based approach was discovered \emph{accidentally} during the hundreds of experiments and trials (163 to be exact) that were conducted in an attempt to train a stable RL agent. It must be noted that this method, though discovered inadvertently, is equivalent to the naive, domain-expert dependent form of the more formal method known as ``Curriculum Learning'' \citep{b:WENG_CURRICULUM,b:NARVEKAR}.

Applying automatic Curriculum Learning requires algorithmic design and implementing complex frameworks \citep{b:PORTELAS}, for example ALP-GMM (absolute learning progress Gaussian mixture model) ``teacher-student'' frameworks. The ``teacher'' neural-network samples parameters from the continuous action space to generate a learning curriculum. Applying automated Curriculum Learning is not readily available as a feature in MATLAB; Graded Learning, on the other hand, requires no programming and can be easily implemented by a control engineer.

\begin{figure}[!ht]
    \centering
    \subfigure[Inexplicable learning curves]{
        \includegraphics[width=0.95\columnwidth]{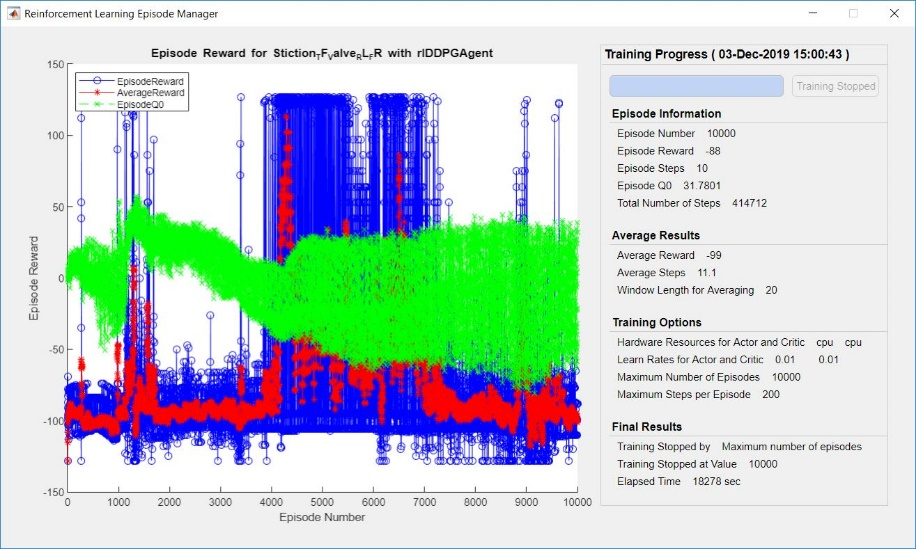}
    }
    \subfigure[Inexplicable controller actions]{
        \includegraphics[width=0.8\columnwidth]{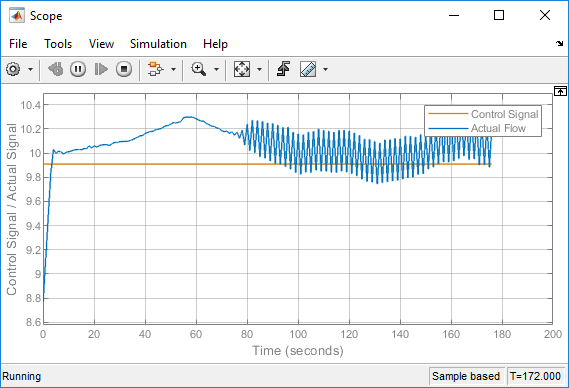}
    }
    \caption{RL agent: Indicative training challenges from our trials}
    \label{fig:GL_Challenges}
\end{figure}

Fig.\ref{fig:GL_Challenges} shows examples of the numerous challenges faced during training, sometimes resulting in experiments with thousands of episodes that did not produce a stable learning curve and sometimes resulting in inexplicable controller actions. Some training trials lasted 20,000 episodes running for over 20 hours and therefore it is important to streamline these efforts. 
 
Graded Learning helped avoid some of these challenges. The intuition for Graded Learning was based on observing how human instructors structure \emph{coaching} of a new skill for apprentices. 

While new skills such as chess or tennis are taught with the final goal in mind, one never starts with the hardest lessons. Foundation level skills are taught first and once some level of proficiency is gained, the student graduates to the next level to handle marginally more complex problems. Skills and experiences gained at any level are retained and progressively built upon while moving to higher levels.

Graded Learning extends this iterative staged approach to RL. The RL task is first broken down to its fundamental level, an agent is trained for $n$ episodes or until convergence criteria is met. Next level of complexity is added to the previous task. \emph{Transfer-learning} is used to ensure previous experiences are retained and built upon. Once this level of task is learned, the process of adding further complexity continues and each time transfer-learning allows to build upon experience gained during the previous levels.

Transfer-learning is a machine learning technique that is used to ``transfer'' the learning i.e. stabilized weights of a neural-network from one task (or domain in general) to another without having to train the neural-network from scratch \citep{b:KARL}. 

The Graded Learning approach was discovered when the time-delay in (\ref{eq:FOPTD}) was reduced to zero and the agent quickly stabilized in contrast to the hundreds of earlier attempts.

\begin{figure*}[p]
    \centering
    \subfigure[Grade-I: $L$=0.1, $f_S$=$\frac{1}{10}\times8.4$, $f_D$=$\frac{1}{10}\times3.524$]{\includegraphics[width=\columnwidth]{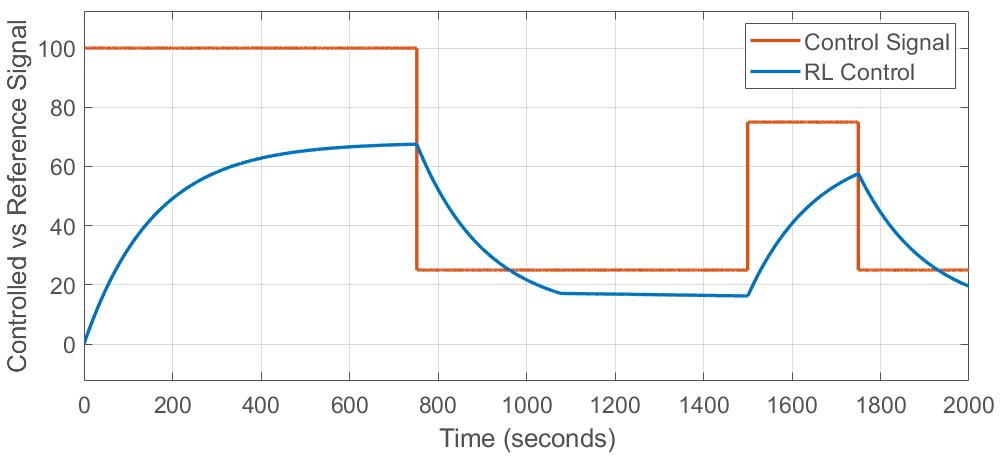}}
    \subfigure[Grade-I.2: \emph{Grade-I} trained for a further 1000 episodes]{\includegraphics[width=\columnwidth]{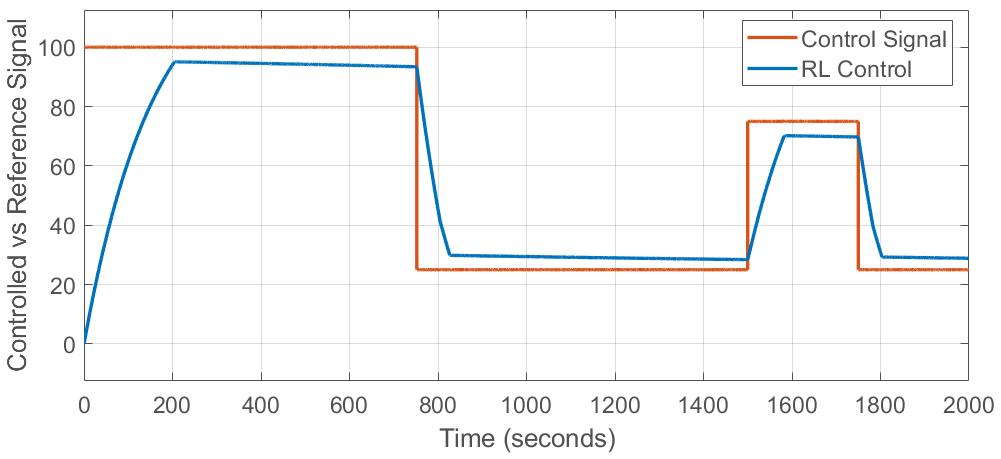}}
    \subfigure[Grade-II: $L$=0.5, $f_S$=$\frac{1}{5}\times8.4$, $f_D$=$\frac{1}{5}\times3.524$]{\includegraphics[width=\columnwidth]{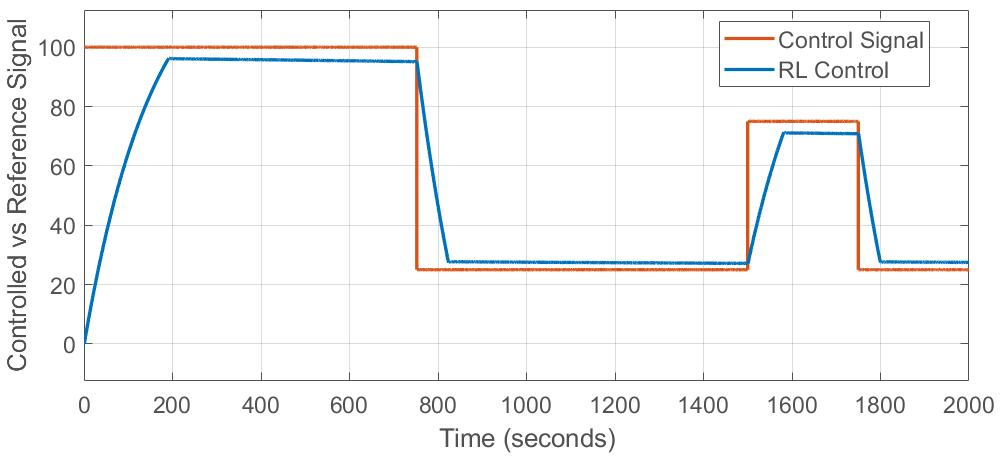}}
    \subfigure[Grade-III: $L$=1.5, $f_S$=$\frac{1}{2}\times8.4$, $f_D$=$\frac{1}{2}\times3.524$]{\includegraphics[width=\columnwidth]{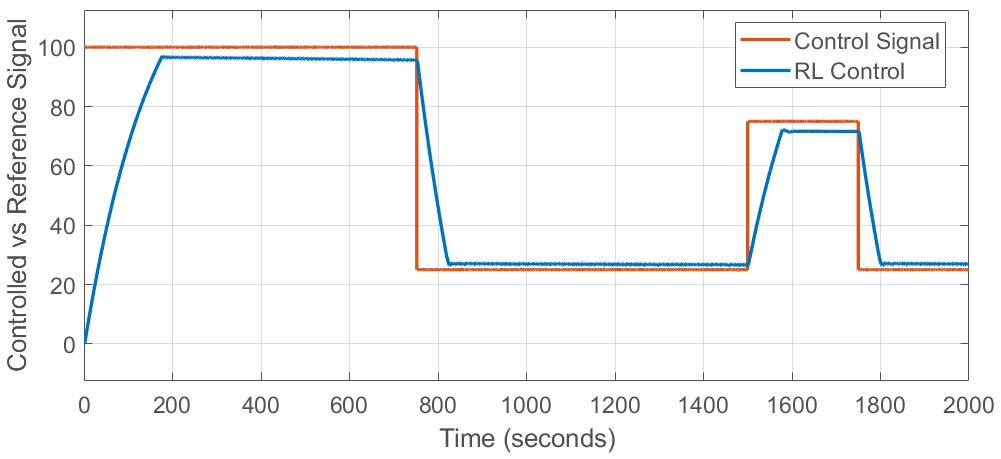}}
    \subfigure[Grade-IV: $L$=1.5, $f_S$=$\frac{2}{3}\times8.4$, $f_D$=$\frac{2}{3}\times3.524$]{\includegraphics[width=\columnwidth]{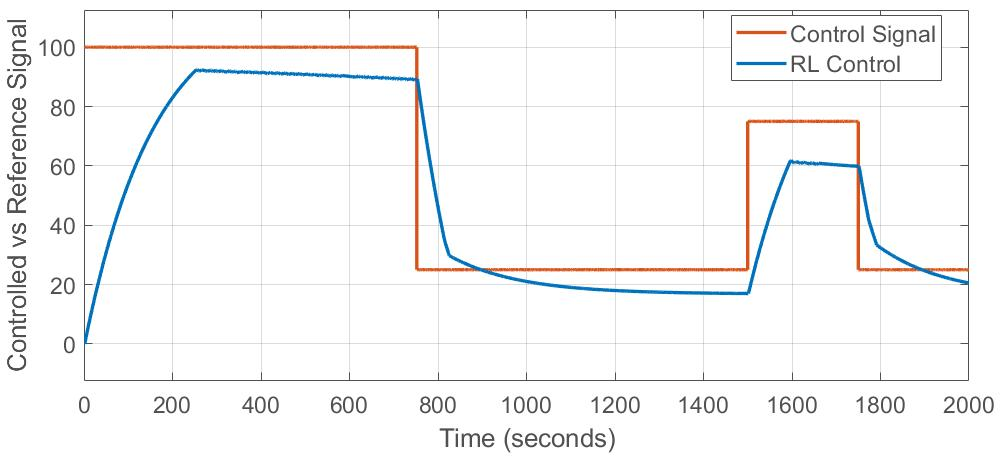}}
    \subfigure[Grade-V: $L$=2.0, $f_S$=$\frac{2}{3}\times8.4$, $f_D$=$\frac{2}{3}\times3.524$]{\includegraphics[width=\columnwidth]{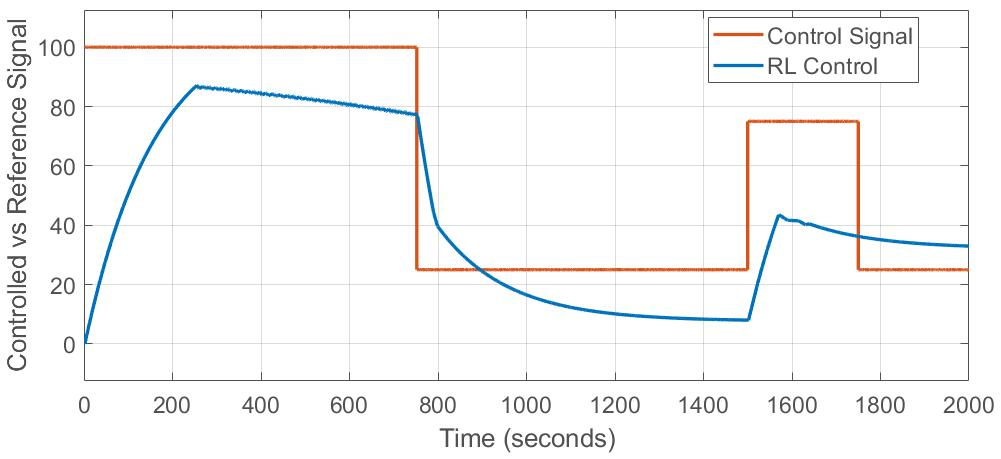}}
    \par\bigskip
    \subfigure[Final learned model: Grade-VI: $L$=2.5, $f_S$=8.4, $f_D$=3.524]{\includegraphics[width=1.4\columnwidth]{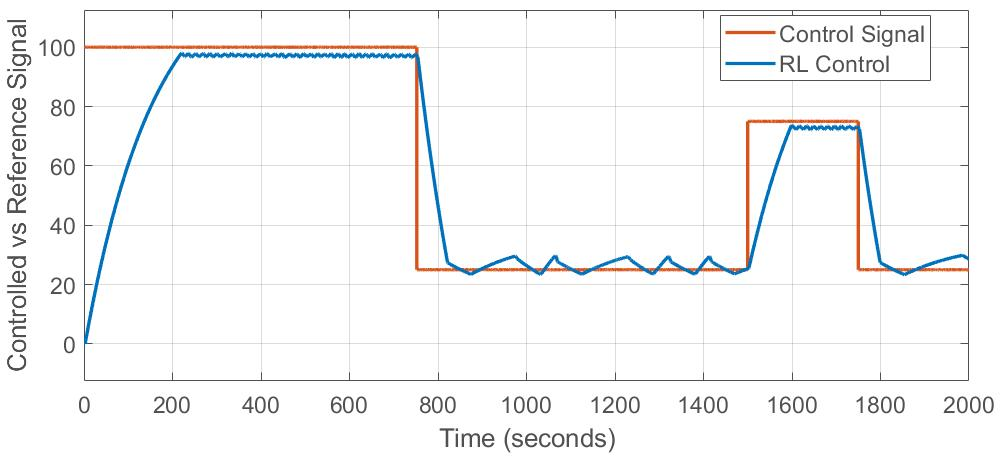}}
    \caption[]{Graded Learning: Progressive learning of task}
    \label{fig:Graded_Learning}
\end{figure*}

Fig.\ref{fig:Graded_Learning} demonstrates the method in action and the agent evolving over six stages of \emph{increasing difficulty}. Parameters that are progressively increased are: time-delay $L$, static friction $f_S$ and dynamic friction $f_D$.

The stability analysis and experimental results presented next, demonstrate that Graded Learning applied to valve control (and possibly other complex industrial systems) appears to be an effective way to coach an RL agent.

\begin{table*}[t]
    \renewcommand{\arraystretch}{1.75}
	\caption{Graded Learning: Staged learning parameters and training times}
    \begin{center}
	\begin{tabular}{|l|r|r|r|r|r|}
	    \hline
		Grade     & $L$ & $f_S$ & $f_D$ & Episodes & Time (h) \\
		\hline
		Grade-I.1 & 0.1 & $\frac{1}{10}\times8.4$  & $\frac{1}{10}\times3.524$  &  930 &  1.67\\ 
		Grade-I.2 & 0.1 & $\frac{1}{10}\times8.4$  & $\frac{1}{10}\times3.524$  & 2000 & 12.35\\ 
		Grade-II  & 0.5 & $\frac{1}{5}\times8.4$   & $\frac{1}{5}\times3.524$   & 1000 &  5.31\\
		Grade-III & 1.5 & $\frac{1}{2}\times8.4$   & $\frac{1}{2}\times3.524$   & 1000 &  5.21\\
		Grade-IV  & 1.5 & $\frac{2}{3}\times8.4$   & $\frac{2}{3}\times3.524$   & 1000 &  4.65\\
		Grade-V   & 2.0 & $\frac{2}{3}\times8.4$   & $\frac{2}{3}\times3.524$   &  500 &  2.27\\
		Grade-VI  & 2.5 & 8.4                      & 3.524                      & 2000 &  7.59\\
		\hline
		\multicolumn{4}{|r|}{Total} & 8430 & 39.05\\
		\hline
	\end{tabular}
	\end{center}	
	\label{table:GL_Settings}
\end{table*}

\section{Experiments, Results and Discussion}
In this section we present the results of experiments conducted to evaluate the RL controller's performance and compare it with the PID (with filter) controller.

Before conducting the experiments a stability analysis of the RL controller must be carried out.

\subsection{Stability Analysis of RL Control}
A basic stability analysis of the RL control is attempted in this section.

\begin{figure}[!ht] 
	\centering 
	\includegraphics[width=\columnwidth]{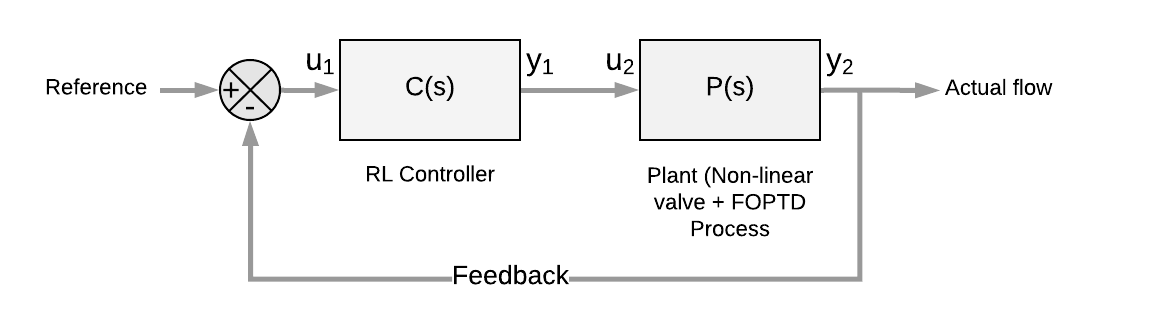} 
	\caption{Block diagram of a single-loop control system}
	\label{fig:SA_block}
\end{figure}

Open-loop transfer-function of the system is $C(s)\cdot P(s)$. Transfer-function of the plant $P(s) = V(s)\cdot G(s)$ where $G(s)$ is the transfer-function of the FOPTD process (\ref{eq:FOPTD}) and $V(s)$ is the transfer-function of the nonlinear valve which is unknown and must be estimated. 

Simulink's Control Design Linearization Analysis\textsuperscript{\texttrademark} tool provides a GUI based interface to generate a \emph{linear approximation} of a nonlinear system, computed across specified input and output points. However, this does not allow any control over the estimation in contrast to MATLAB's \texttt{tfest} function.

The programmatic method allows a user to estimate the transfer-function by specifying the number of poles (\texttt{np}) and zeros (\texttt{nz}). Additionally the \texttt{iodelay} parameter allows experimenting the effect of time-delays in physical systems. This MATLAB function is based on \citep{b:GARNIER}.

\begin{verbatim}
	sys = tfest(data, np, nz, iodelay)
\end{verbatim}

\begin{figure}[!ht] 
	\centering 
	\includegraphics[width=\columnwidth]{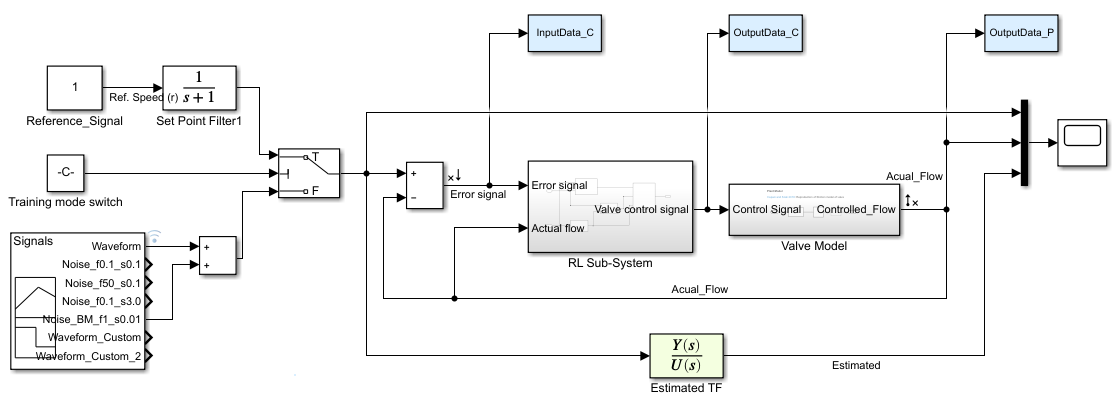} 
	\caption{Setup for transfer-function estimation}
	\label{fig:SA_setup}
\end{figure}

The block-diagram Fig.\ref{fig:SA_block} shows the points at which data $u_1$ and $y_1$ will be tapped to estimate the controller transfer-function $C(s)$ and points $u_2$ and $y_2$ to estimate the complete plant transfer-function $P(s)$. Fig.\ref{fig:SA_setup} is the Simulink setup to assist in estimation. 
 
\smallskip
\textbf{Estimated transfer-functions}:
The estimated continuous-time transfer-function for the plant is (\ref{eq:TF_plant}), with a fit of $97.15\%$ and MSE of $0.7921$, while that of the controller is (\ref{eq:TF_controller}).

\begin{equation}
	\frac {0.002255 s^2 - 1.904\times 10^{-5} s + 8.563\times 10^{-7}} {s^3 + 0.01305 s^2 + 9.451\times 10^{-5} s + 2.278\times 10^{-7}}
 \label{eq:TF_plant}
\end{equation}

\begin{equation}
    \frac {0.09455 s^2 + 0.0005729 s + 1.609\times 10^{-6}} {s^3 + 0.2312 s^2 + 0.001939 s + 1.195\times 10^{-7}}
	\label{eq:TF_controller}
\end{equation}

\begin{figure}[!ht] 
	\centering 
	\includegraphics[width=\columnwidth]{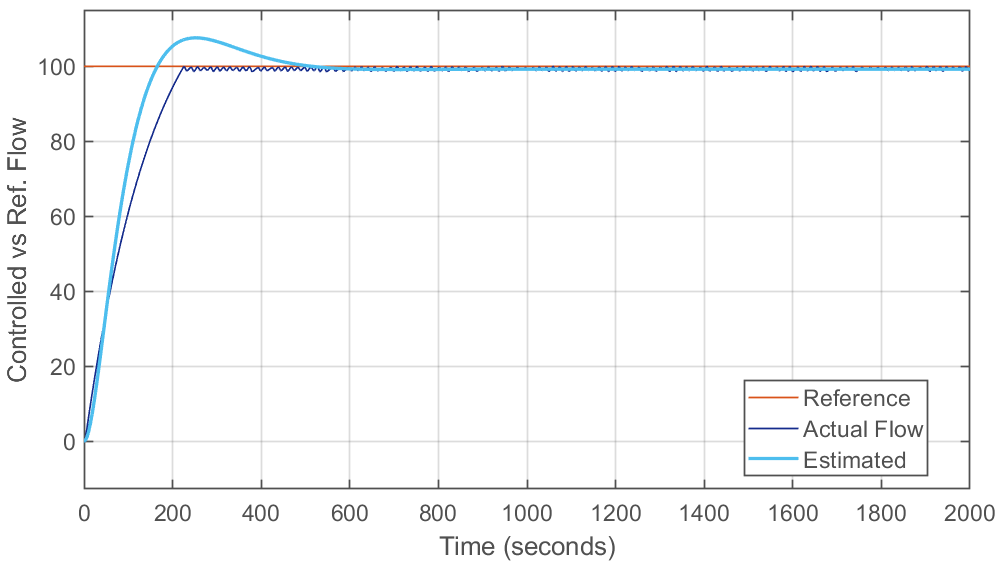} 
	\caption{Stability analysis: Comparing plant responses using the estimated transfer-functions}
	\label{fig:SA_EstimatedSignal}
\end{figure}

We plot (Fig.\ref{fig:SA_EstimatedSignal}), the plant's response using the estimated transfer-functions against the actual response; to ensure that the estimations are reasonable for conducting a basic stability analysis.

\smallskip

\begin{figure}[!ht] 
	\centering 
	\includegraphics[width=\columnwidth]{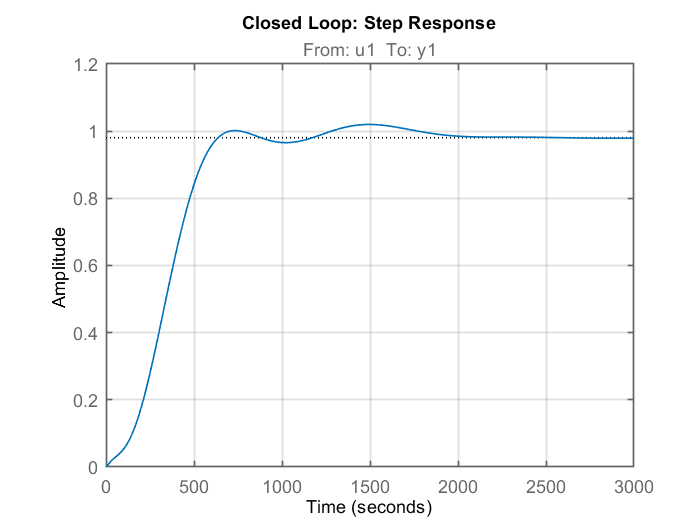} 
	\caption{RL controller: Step response}
	\label{fig:SA_StepResponse}
\end{figure}

\begin{figure}[!ht] 
	\centering 
	\includegraphics[width=\columnwidth]{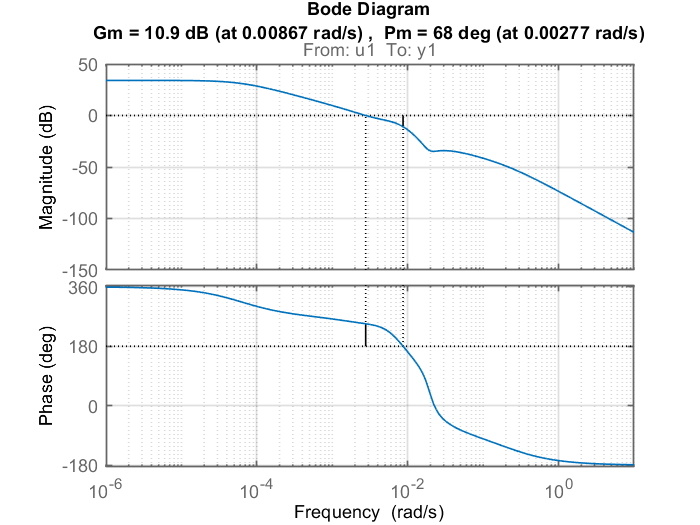} 
	\caption{RL controller: Open-loop Bode loop}
	\label{fig:SA_OpenLoop_BodePlot}
\end{figure}

\textbf{Stability analysis}:
The step-response in Fig.\ref{fig:SA_StepResponse} shows a stable closed-loop system. The open-loop Bode plot, Fig.\ref{fig:SA_OpenLoop_BodePlot}, shows a gain-margin of 10.9 dB and a phase-margin of 68.0 degrees, indicating a fairly stable system. 
  
\subsection{Experiments and Results}
In this section we present the results of experiments conducted on a unified framework that tests two valve control strategies --- PID (with filter) and DDPG RL. A critical time-domain analysis of experiments with varying control signals, varying noise strengths and disturbance points and effect of a plant with process-loop perturbations is presented.\\

\textbf{Experiments conducted}:
\begin{enumerate}
    \item Arbitrarily assumed constant reference level.
    \item Benchmark waveform (with noise).
    \item Benchmark waveform subject to disturbances at:
    \begin{itemize}
        \item  Controller input (i.e. reference signal).
        \item  Plant input (i.e. controlled signal fed to plant).
        \item  Plant output (i.e. system output).
    \end{itemize}
    \item Practical example of a ``water-supply'' valve, subject to ground-borne vibrations of passing trains.
    \item Plant experiencing process loop-perturbations.
    \item Arbitrary control waveform.
\end{enumerate}

\subsubsection{Experiment-1: Constant reference signal}\smallskip
\textbf{Experiment}: A basic analysis is best done on a simple constant reference flow rate arbitrarily set at 100 and run over 2,000 $s$. Reference signal is superimposed with benchmark Gaussian noise ($\mu=0.0, \sigma=0.01$).

\begin{figure}[!ht]
    \centering \includegraphics[width=\columnwidth]{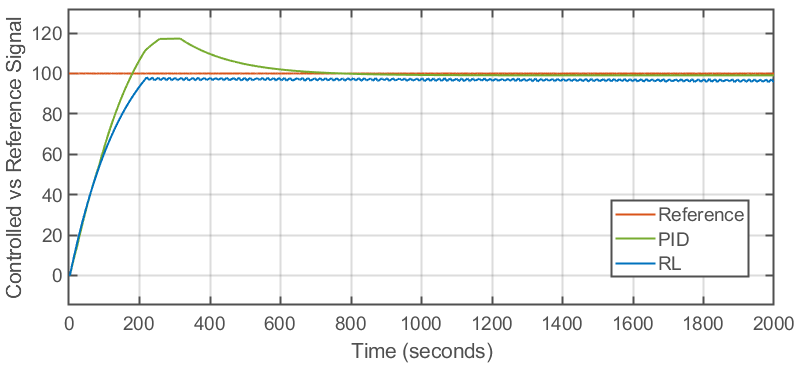}
    \caption{Expt.-1: Constant reference signal}
	\label{fig:Expt_1}
\end{figure}

\textbf{Observations}: In Fig.\ref{fig:Expt_1} the PID shows a large overshoot and settles in about 700 $s$. The RL strategy demonstrates close to ideal damping and a quicker settling time of about 220 $s$. The RL trajectory shows tiny ripples against the PID's smoother profile. These oscillations can reduce the remaining-useful-life (RUL) of a mechanical system and we study this by conducting a \emph(simplified) two factor DOE (design of experiments).

We vary the two factors; time-delay and valve friction (combined static and dynamic) as shown in Table \ref{table:DoE_levels}. The default values of time-delay $L = 2.5$, static-friction $f_S=8.40$ and $f_D=3.524$ are treated as the high-levels and we lower each by a factor of 100 to obtain the low-levels as shown in Table \ref{table:DoE_values}.

\begin{table}[H]
	\caption{DoE table}
    \renewcommand{\arraystretch}{1.5}
    \begin{center}
	\begin{tabular}{ |c|c| } 
		\hline
		Time-delay ($L$) & Friction values ($f_S$, $f_D$)\\
		\hline
		Low  & Low  \\
		Low  & High \\
		High & Low  \\
		High & High \\
		\hline
	\end{tabular}
	\end{center}	
	\label{table:DoE_levels}
\end{table}

\begin{table}[H]
	\caption{DoE table with actual values}
    \renewcommand{\arraystretch}{1.5}
    \begin{center}
	\begin{tabular}{ |c|c|c| }
		\hline
		$L$ & $f_S$ & $f_D$ \\
		\hline
		0.025 & 0.084 & 0.0352 \\ 
		0.025 & 8.400 & 3.524 \\ 
		2.500 & 0.084 & 0.0352 \\ 
		2.500 & 8.400 & 3.524 \\ 
		\hline
	\end{tabular}
	\end{center}
	
	\label{table:DoE_values}
\end{table}

\begin{figure}[!ht]
    \centering
    \subfigure[$L$=Low; $f_S$ and $f_D$=Low]{
        \includegraphics[width=0.45\columnwidth]{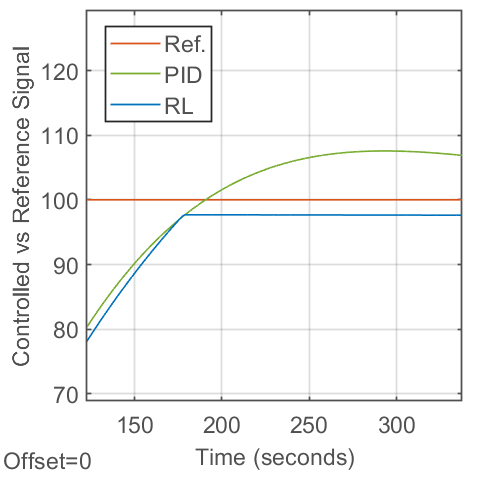}
        \label{fig:Expt_1a}
    }
    \subfigure[$L$=Low; $f_S$ and $f_D$=High]{
        \includegraphics[width=0.45\columnwidth]{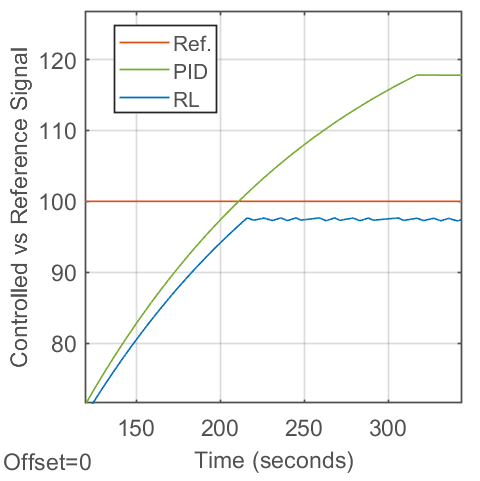}
        \label{fig:Expt_1b}
    }
	\subfigure[$L$=High; $f_S$ and $f_D$=Low]{
        \includegraphics[width=0.45\columnwidth]{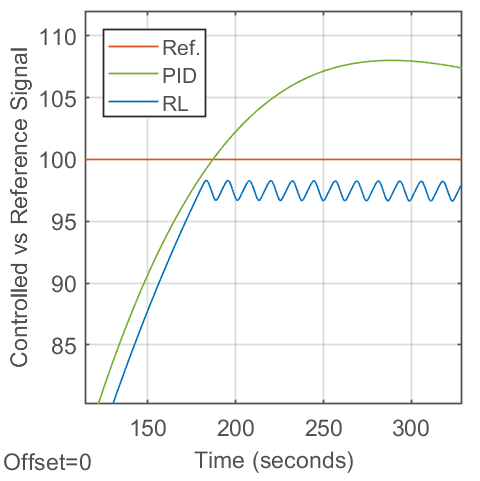}
        \label{fig:Expt_1c}
    }
	\subfigure[$L$=High; $f_S$ and $f_D$=High]{
        \includegraphics[width=0.45\columnwidth]{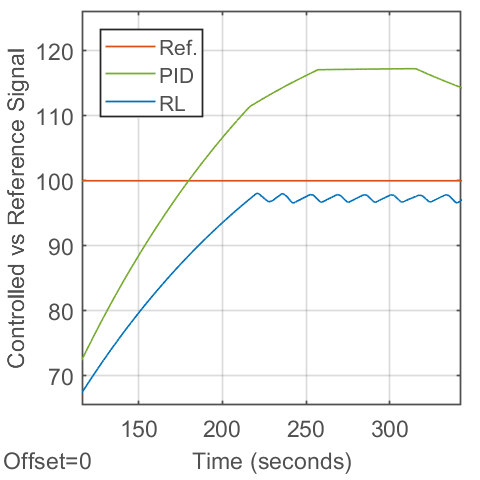}
        \label{fig:Expt_1d}
    }
    \caption{Expt.-1: DoE with time-delay and friction parameters}
    \label{fig:Expt1_subfigures}
\end{figure}

Fig.\ref{fig:Expt_1a} highlights the RL's capability to produce a very smooth profile when both the factors are low. This implies that the oscillations are not introduced by the RL technique. Fig.\ref{fig:Expt_1c} shows that the cause of oscillatory behavior is mainly due to the time-delay factor.

While the PID strategy (\ref{eq:pid_practical}), is implemented with a filter that suppresses noise, \emph{no} filters were added to the RL setup to better understand the natural response of RL control strategies.

\subsubsection{Experiment-2: The benchmark signal}
\begin{figure}[!ht]
    \centering \includegraphics[width=\columnwidth]{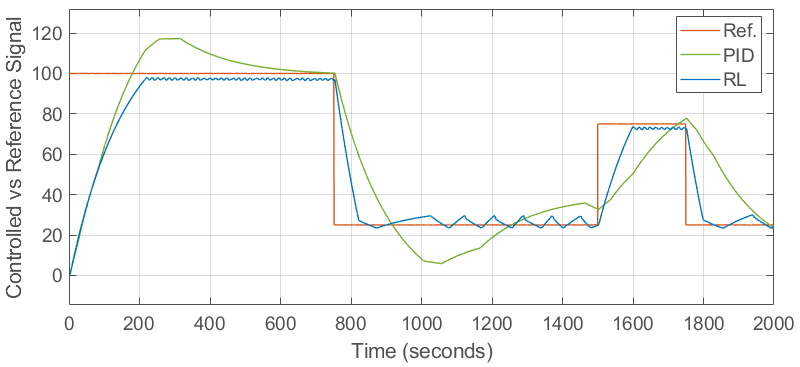} 
    \caption{Expt.-2: Benchmark waveform} 
	\label{fig:Expt_2}
\end{figure}

\begin{figure}[!ht]
    \centering
    \subfigure[Zoomed section 1]{
        \includegraphics[width=0.45\columnwidth]{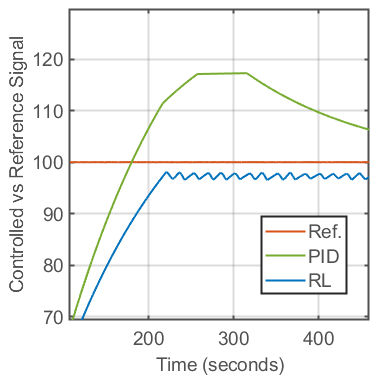}
        \label{fig:Expt_2a}
    }
    \subfigure[Zoomed section 2]{
        \includegraphics[width=0.45\columnwidth]{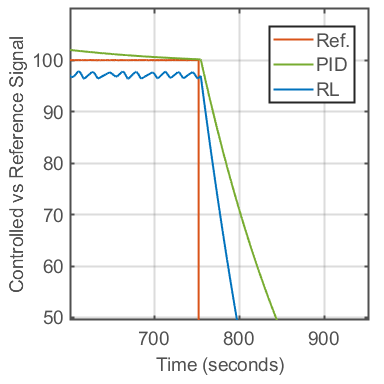}
        \label{fig:Expt_2b}
    }
	\subfigure[Zoomed section 3]{
        \includegraphics[width=0.45\columnwidth]{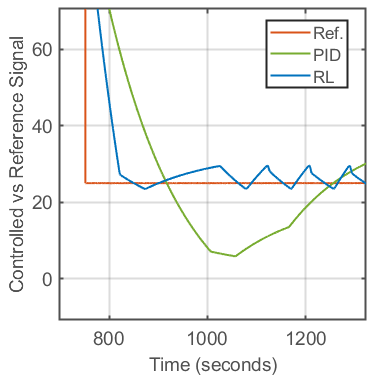}
        \label{fig:Expt_2c}
    }
	\subfigure[Zoomed section 4]{
        \includegraphics[width=0.45\columnwidth]{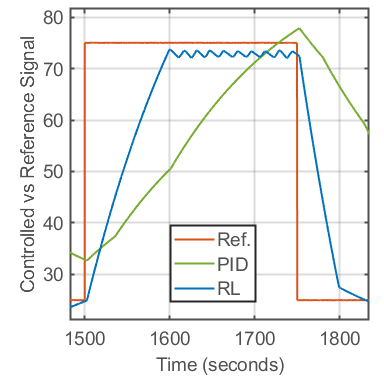}
        \label{fig:Expt_2d}
    }
    \caption{Expt.-2: Zoomed sections of the benchmark signal}
    \label{fig:Expt2_subfigures}
\end{figure}

\smallskip
\textbf{Experiment}: The waveform profile used in \citep{b:CAPACI}, with Gaussian noise ($\mu=0.0, \sigma=0.01$), is subject to both strategies. We also zoom sections of time-domain plot Fig.\ref{fig:Expt_2} and observe them more closely in Fig.\ref{fig:Expt2_subfigures}. 

It is observed that PID shows higher over- and under-shoots. If such a valve controls fluid flow, the higher and lower fluid quantities could be detrimental to the product quality. In \ref{fig:Expt_2} the shifted PID waveform after 800 $s$ could be detrimental to the process if it depends on the timing of the flow of fluid. In contrast RL control shows better tracking of reference signal. 

\subsubsection{Experiment-3.a: Noise at controller input}\smallskip
\textbf{Experiment}: Increase noise at the controller input ($\mu=0, \sigma=3.0, 1 Hz$)

\begin{figure}[!ht]
    \centering
    \subfigure[Entire trajectory plot]{
        \includegraphics[width=\columnwidth]{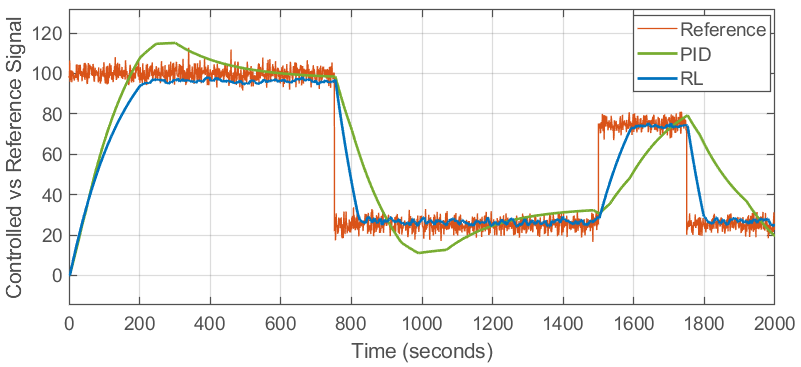}
        \label{fig:Expt_3a}
    }
    \subfigure[Zoomed section]{
        \includegraphics[width=0.5\columnwidth]{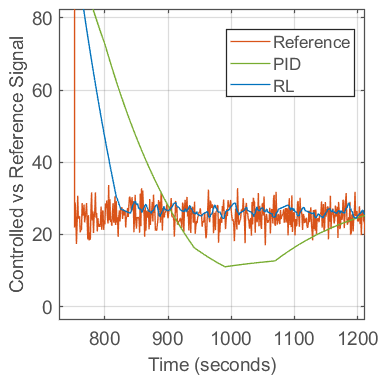}
        \label{fig:Expt_3a_zoomed}
    }
    \caption{Expt.-3.a: Noise at controller input ($\mu=0, \sigma=3.0, 1 Hz$)}
\end{figure}

\textbf{Observations}: Fig.\ref{fig:Expt_3a} and \ref{fig:Expt_3a_zoomed} show almost no impact on the PID when compared with Experiment-2 (lower noise at input) but increased impact on the RL trajectory, demonstrating the PID strategy's superior noise attenuation capabilities. The RL continues to closely track the reference signal (\emph{along} with the noise).  

\subsubsection{Experiment-3.b: Noise at plant input}\smallskip
\textbf{Experiment}: Shift the source of noise to the plant input ($\mu=0, \sigma=3.0, 1 Hz$).

\begin{figure}[!ht]
    \centering \includegraphics[width=\columnwidth]{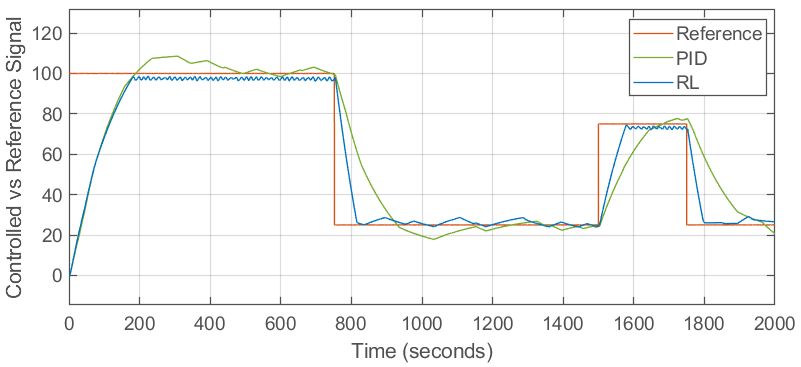} 
    \caption{Expt.-3.b: Noise at plant input ($\mu=0, \sigma=3.0, 1 Hz$)}
	\label {fig:Expt_3b}
\end{figure}

\textbf{Observations}: Fig.\ref{fig:Expt_3b} shows that the PID trajectory is now impacted and it looses its relatively smooth output seen in Experiment-1 and Experiment-2. The RL strategy on the other hand remains \emph{unaffected} when compared to Experiment-1. The PID strategy adjusts itself based on the error signal and hence shows a change in behavior while RL strategy does not.

\subsubsection{Experiment-3.c: Noise at plant output}\smallskip
\textbf{Experiment}: Effect of noise at the plant output ($\mu=0, \sigma=3.0, 1 Hz$).

\begin{figure}[!ht]
    \centering
    \subfigure[Entire trajectory plot]{
        \includegraphics[width=\columnwidth]{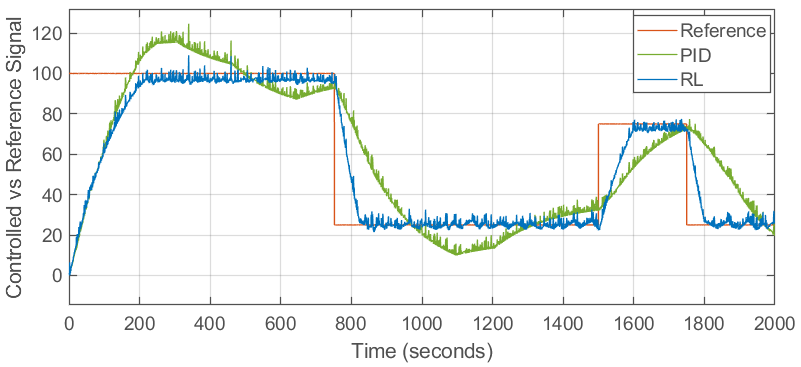}
        \label{fig:Expt_3c_full}
    }
    \subfigure[Zoomed section]{
        \includegraphics[width=0.5\columnwidth]{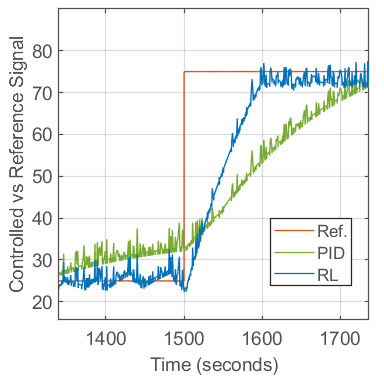}
        \label{fig:Expt_3c_zoomed}
    }
    \caption{Expt-3c: Noise at plant output ($\mu=0, \sigma=3.0, 1 Hz$)}
    \label{fig:Expt3c}
\end{figure}

\textbf{Observations}: Fig.\ref{fig:Expt_3c_full} shows that both RL and PID strategies are affected equally.

\subsubsection{Experiment-4: Water-supply valve, subject to ground-borne vibrations}\smallskip
\textbf{Experiment}: Valve applications could often be exposed to extremely harsh conditions. A water-supply system, for example, may face ground-borne vibrations such as from passing railways, that is in the range of about 30--200 Hz and varying amplitudes \citep{b:TRAIN}. Since the control-valve assembly will often be placed in shielded environments frequencies between 30--100 Hz were assumed for simulation.

\begin{figure}[!ht]
    \centering
    \subfigure[Entire trajectory plot]{
        \includegraphics[width=\columnwidth]{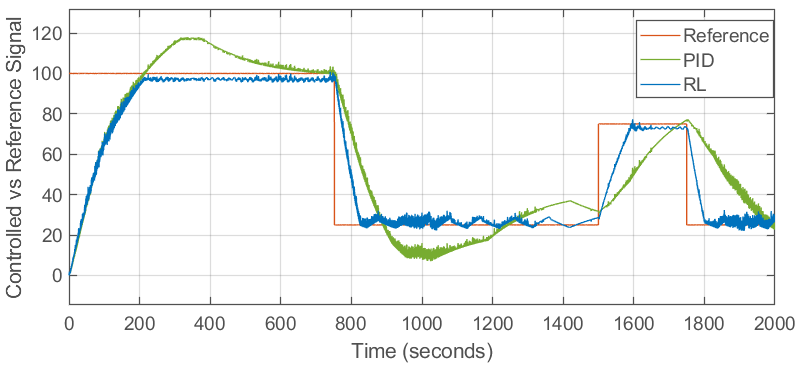}
        \label{fig:Expt_4_full}
    }
    \subfigure[Zoomed section]{
        \includegraphics[width=0.5\columnwidth]{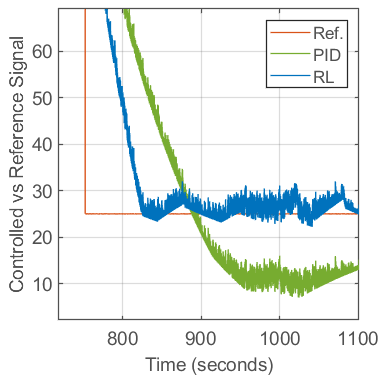}
        \label{fig:Expt_4_zoomed}
    }
    \caption{Expt.-4: Ground-borne vibrations of a passing train}
    \label{fig:Expt_4}
\end{figure}

\textbf{Observations}: Figures \ref{fig:Expt_4_full} and \ref{fig:Expt_4_zoomed} show that, as in Experiment-3.c, the impact of noise is similar on both strategies and RL continues to track the reference signal better than PID.

\subsubsection{Experiment-5: Arbitrary control waveform with benchmark noise signal}\smallskip
\textbf{Experiment}: Test the generalization capability of RL \emph{training} strategy vis-à-vis generalization of PID tuning. ``Training'' signal for both the strategies was the benchmark waveform and this experiment subjected them to a completely different waveform.

\begin{figure}[!ht]
    \centering
	\subfigure[Arbitrary control waveform]{
        \includegraphics[width=\columnwidth]{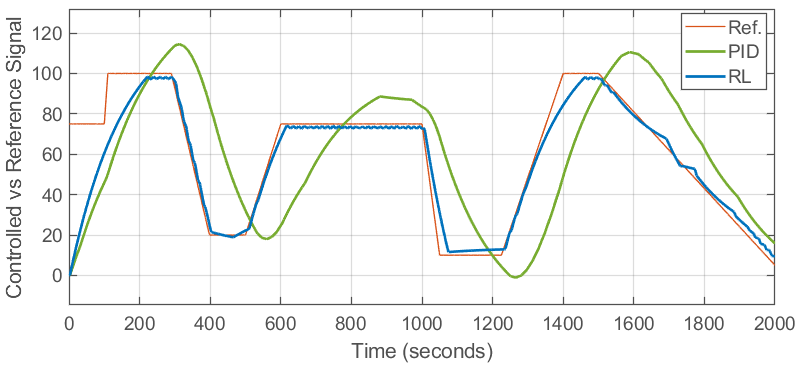}
        \label{fig:Expt_5_full}
    }
    \subfigure[Zoomed section 1]{
        \includegraphics[width=0.45\columnwidth]{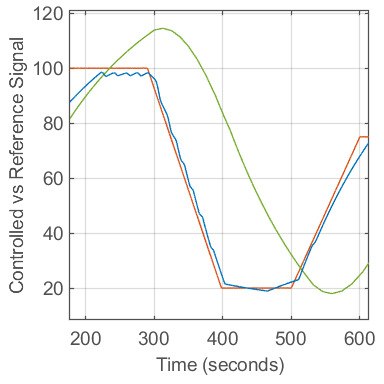}
        \label{fig:Expt_5_zoomed_1}
    }
    \subfigure[Zoomed section 2]{
        \includegraphics[width=0.45\columnwidth]{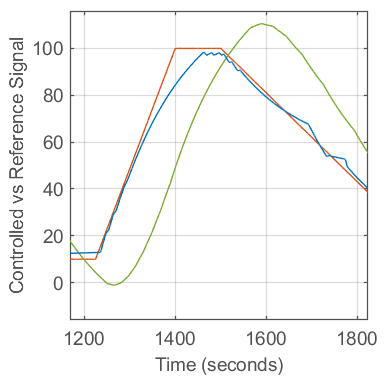}
        \label{fig:Expt_5_zoomed_2}
    }
    \caption{Expt.-5: Response to an arbitrary control waveform}
    \label{fig:Expt_5}
\end{figure}

\textbf{Observations}: Fig.\ref{fig:Expt_5} shows that the RL controller out-performs the PID strategy considerably in this experiment. The RL controller tracks the arbitrary reference more closely and this demonstrates the importance of the training strategy in effective generalization. The PID trajectory, on the other hand shows a significant lag while tracking the reference and if such a valve controls fluid flow, the untimely higher or lower fluid quantities could be detrimental to the product quality.

Small ripples are evident in sections of the RL controlled trajectory.

\subsubsection{Experiment-6: Benchmark plant with process loop-perturbations}\smallskip
\textbf{Experiment}: Evaluate resistance to severe process-loop perturbations modeled as a $3^{rd}$ order transfer-function (\ref{eq:perturbation}) \citep{b:CAPACI}.

\begin{equation}
	G(s) = \frac {1} {(1s+1)(5s+1)(10s+1)}
	\label {eq:perturbation}
\end{equation}

\begin{figure}[!ht]
    \centering
    \subfigure[Response to benchmark signal]{
        \includegraphics[width=\columnwidth]{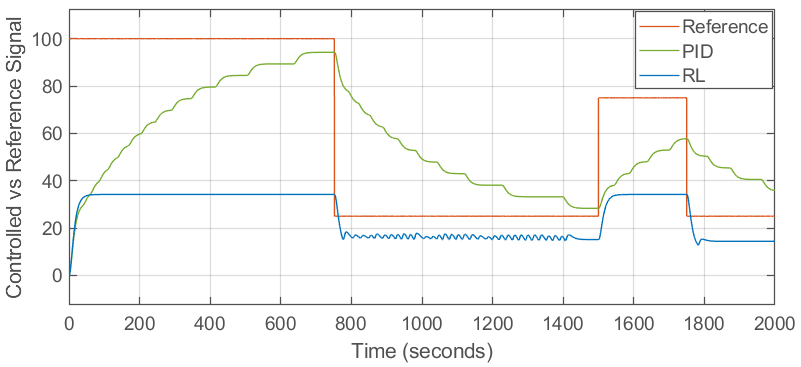}
        \label{fig:perturbation_100}
    }
    \subfigure[Response to benchmark signal with lower strength]{
        \includegraphics[width=\columnwidth]{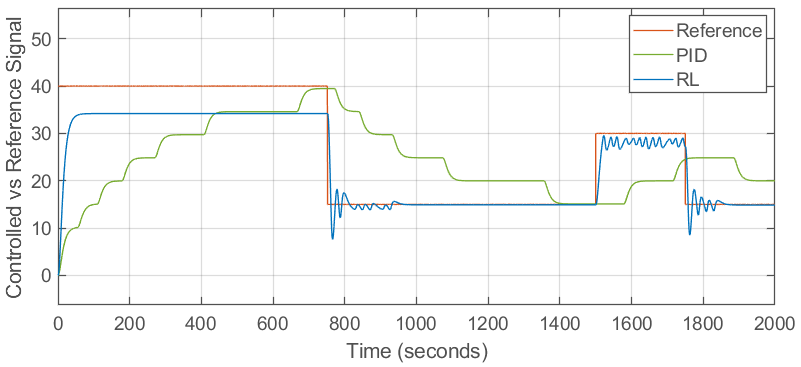}
        \label{fig:perturbation_40}
    }
    \caption{Expt.-6: Response to plant with perturbations}
    \label{fig:Expt_6}
\end{figure}

\textbf{Observations}: A severe \emph{limitation} of the RL controller is evident in this experiment. Fig.\ref{fig:Expt_6} shows a significantly stunted output, clamped smoothly at around 35.0. The setup was then tested on a lower magnitude reference Fig.\ref{fig:perturbation_40} and the RL continues to be clamped at the same level, 35.0. PID seems to scale to different levels under the influence of perturbations albeit with significant error. The RL controller shows increased oscillatory behavior at the lower flow level.

\subsection{Discussion: Experiential Learning Validated against Published Research}\smallskip
\noindent
Merriam-Webster: ``Experiential: relating to, \emph{derived from}, or providing \emph{experience}"\\

A total of 163 experiments were conducted during this research. When experiments did not respond to seemingly logical steps it led to severe frustration. In a quest to find answers for some of the strange observations, research was conducted to relate these to previously published studies and it highlighted the several known challenges that exist; reminding one that RL is still an emerging field.

Early adopters of RL for control are encouraged to try both the Graded Learning method and study the literature referenced in this section --- which is a collection of studies conducted at Google, MIT and Berkeley; \citep{b:SONG, b:HENDERSON, b:HARDT, b:ZHANG}.

In \citep{b:HENDERSON}, effects of hyperparameters and their tuning are analyzed with respect to network-architecture, rewards scaling and reproducibility on model-free, policy-gradient based algorithms for continuous control and is therefore directly applicable to the subject of this paper. 

\subsubsection{Over-fitting and saturation}
\begin{figure}[htbp]
	\centering
    \subfigure[Example-1]{\includegraphics[width=0.95\columnwidth]{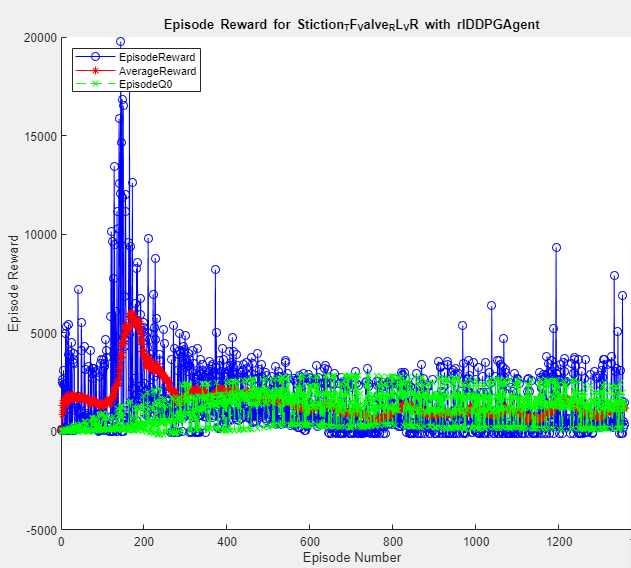}}
	\subfigure[Example-2]{\includegraphics[width=0.95\columnwidth]{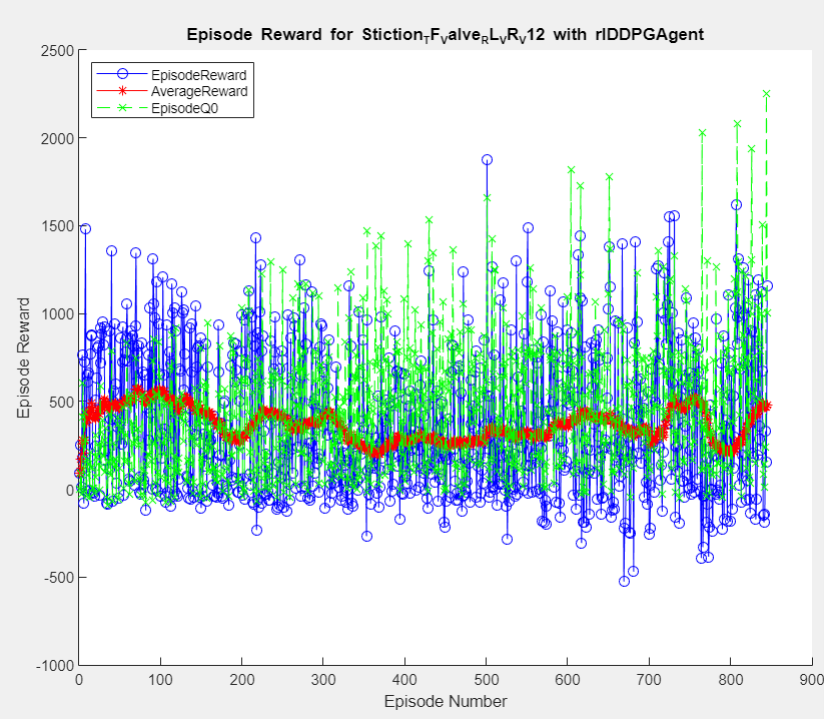}}
    \caption{Examples from our trials showing effects of over-training and network saturation}
    \label{fig:Saturation}
\end{figure}

For physical systems, there is always an upper limit of performance that the agent cannot surpass. However this is not known before hand and one often pushes the agent to continue training for hours. Significant neural-network saturation was observed in several of the training attempts, Fig.\ref{fig:Saturation}.

Over-fitting in RL is being studied only recently. \citeauthor{b:SONG} (\citeyear{b:SONG}) observe that in model-free RL, the agent often mistakenly correlates rewards with spurious observation-space features (``observational overfitting"). In particular they have studied over-fitting with linear quadratic regulators (LQR) using neural-networks and show that under Gaussian initialization of the policy using gradient descent, a generalization gap ``must necessarily exist''.

\citeauthor{b:HARDT} (\citeyear{b:HARDT}) provides a theoretical proof, that stochastic-gradient methods employing parametric models, when trained using \emph{fewer} iterations, have vanishing generalization errors. They argue this by experiments conducted and using stability criteria established for learning algorithms devised by \citeauthor{b:BOUSQUET}. They conclude, that shortened training time by itself, sufficiently prevents over-fitting. This paper is important for extending the stability criteria developed for supervised learning to iterative algorithms, such as RL \citep{b:BOUSQUET}.

\subsubsection{Sensitivity to network architecture}
Four policy-gradient methods including the DDPG are analyzed in \citep{b:HENDERSON}. While ReLU activations were stated to perform best, the effects were not consistent across algorithms or hyperparameter settings.

\subsubsection{Sensitivity to reward-scaling}
A large and sparse reward scale causes network saturation; resulting in inefficient learning. Reward rescaling is a technique recommended to improve results for DDPG. This is achieved by multiplying by a scalar such as 0.1 or clipping to \texttt{[0, 1]} \citep{b:DUAN}.

\begin{figure}[htbp] 
    \centering
    \subfigure[Published results \citep{b:HENDERSON}]{
        \includegraphics[width=0.95\columnwidth]{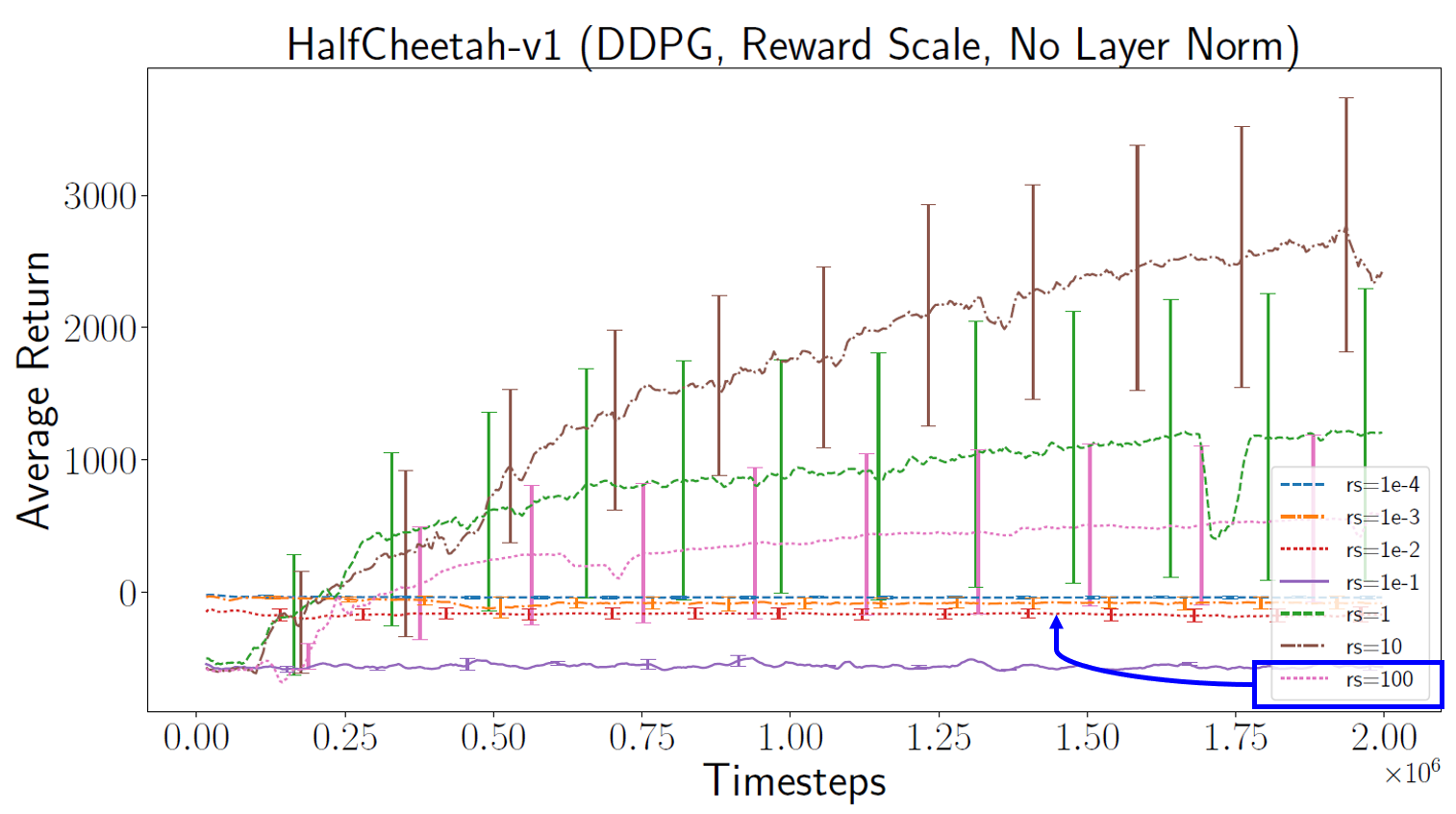}
        \label{fig:Reward_scaling_Henderson}
    }
    \subfigure[Inefficient learning at $10^2$ reward scales]{
        \includegraphics[width=0.95\columnwidth]{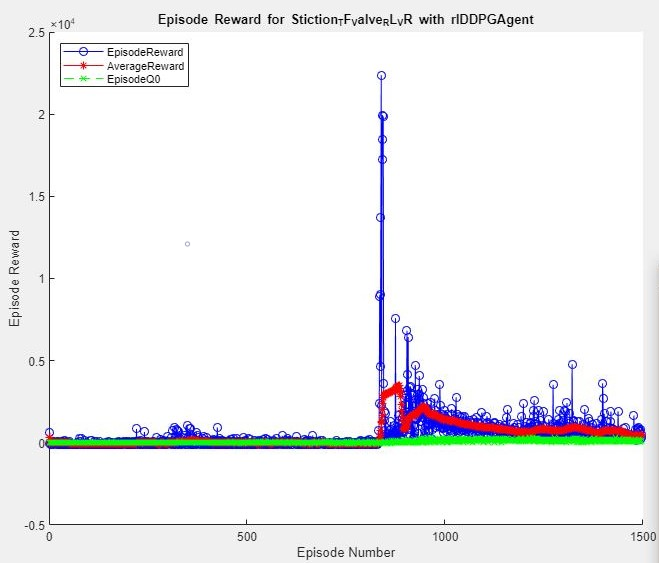}
        \label{fig:Reward_scaling_observed}
    }
    \caption{Effect of large reward spaces}
    \label{fig:Reward_scaling}
\end{figure}

Fig.\ref{fig:Reward_scaling_Henderson} shows the effects of different reward scales conducted by \citeauthor{b:HENDERSON} (\citeyear{b:HENDERSON}); with $10^2$ highlighted for comparison with similar scales, observed in trials we conducted (Fig.\ref{fig:Reward_scaling_observed}). Reward scale can exhibit a large impact and must be given due consideration \citep{b:HENDERSON}. 
  
\subsubsection{Sensitivity to noise parameter}
DDPG uses the Ornstein-Uhlenbeck process to aid exploration. The effect of noise hyperparameter was not very easily ascertainable.

\begin{figure}[htbp]
    \centering
    \subfigure[Variance decay-rate=$3\times10^{-5}$. Exploration decay after 150 episodes]{
        \includegraphics[width=0.95\columnwidth]{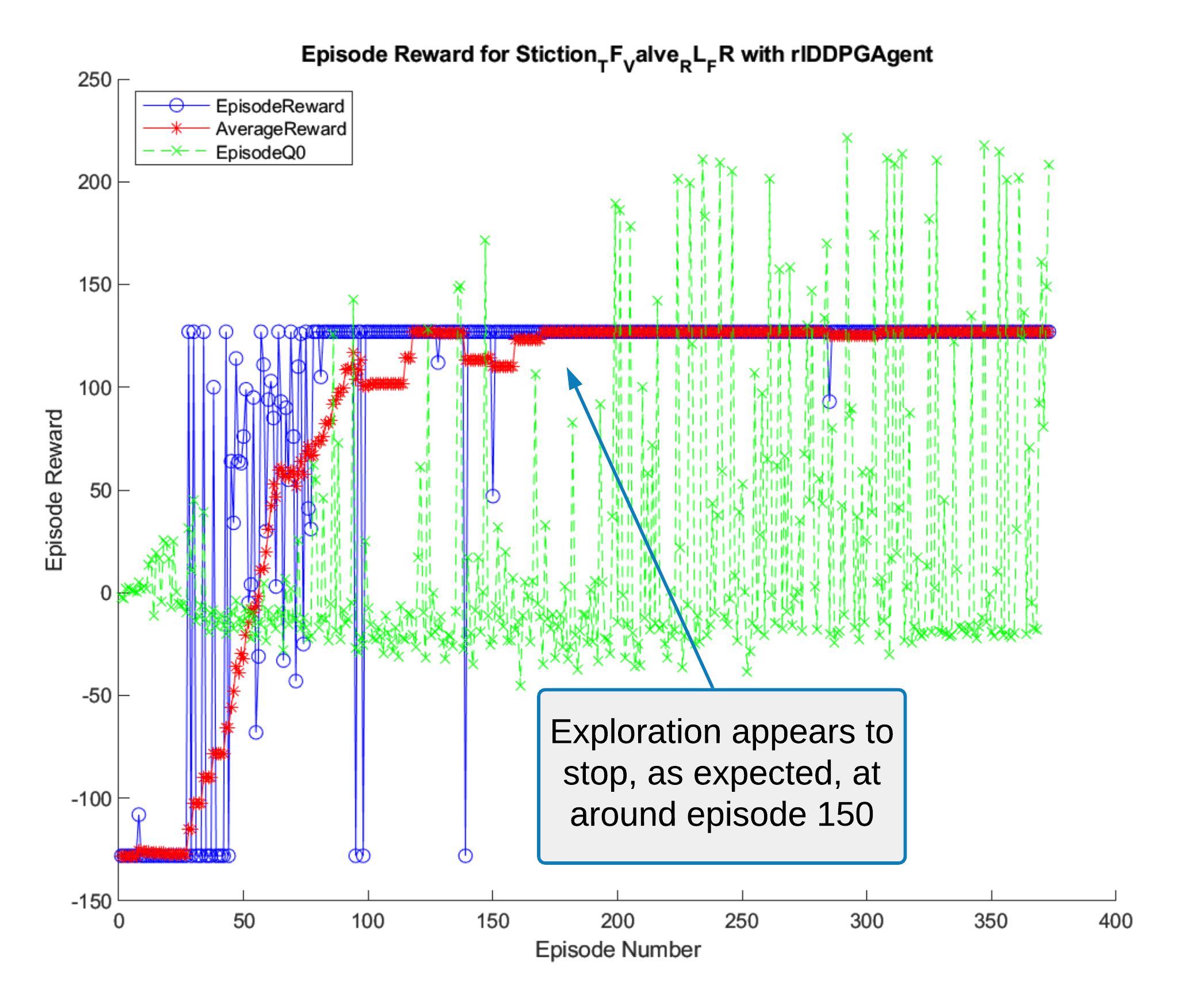}
        \label{fig:OUP_ep350}
    }
    \subfigure[Variance decay-rate=$3\times10^{-5}$. Exploration explosion after 650 episodes]{
        \includegraphics[width=0.95\columnwidth]{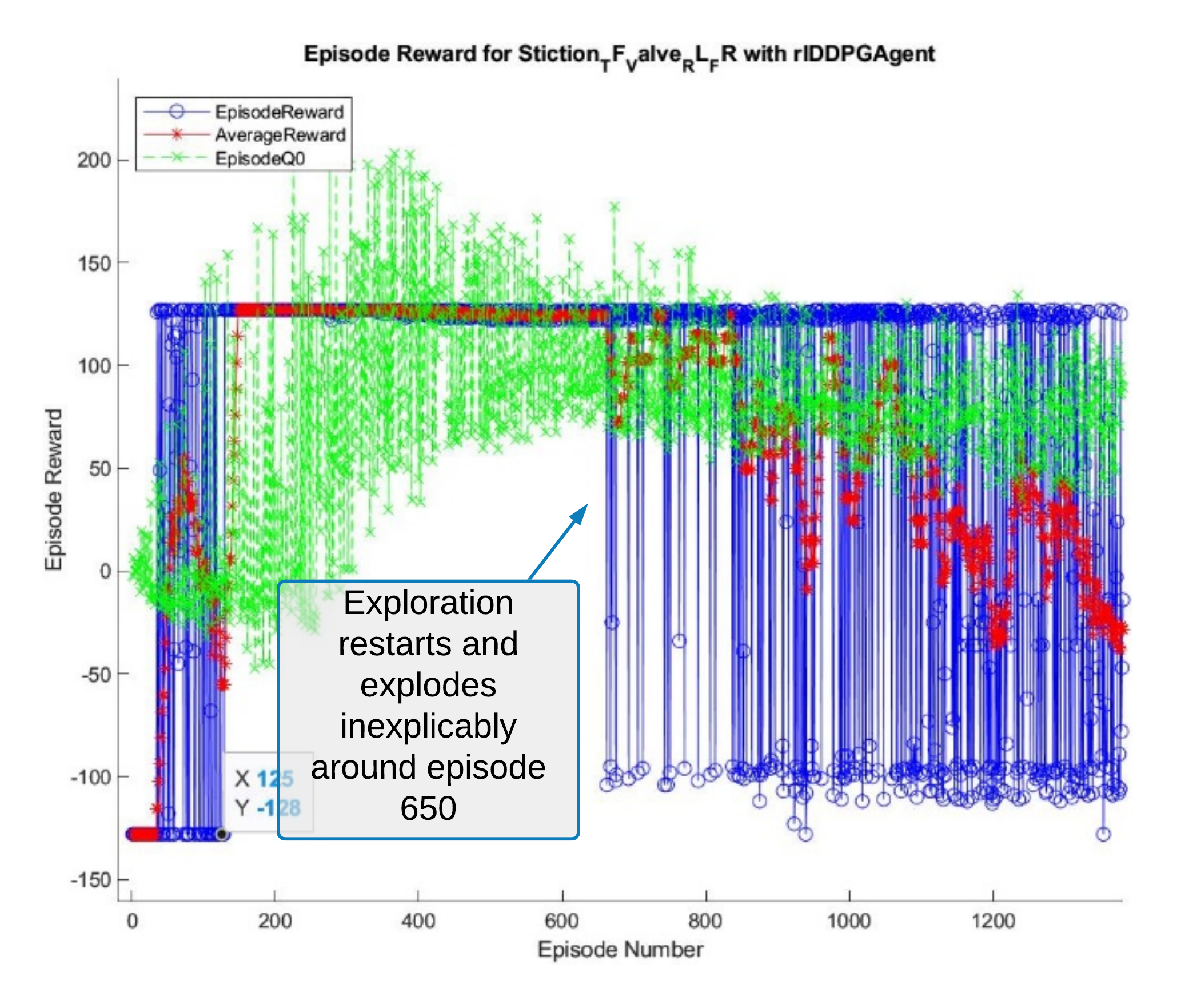}
        \label{fig:OUP_ep1300}
    }
    \caption{Effect of OUP parameters on exploration}
    \label{fig:OUP_sensitivity}
\end{figure}

\begin{figure}[htbp]
    \centering
    \includegraphics[width=0.95\columnwidth]{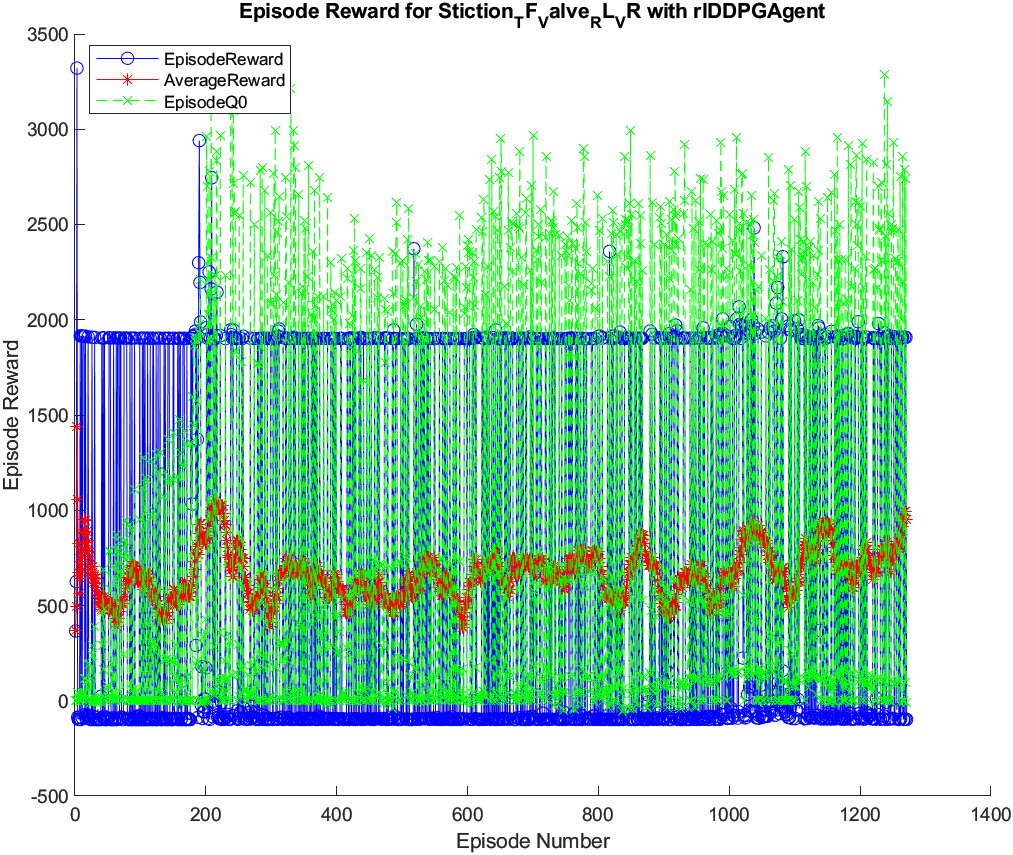}
    \caption{No signs of exploration decay for a decay-rate=$3\times10^{-4}$}
    \label{fig:OUP_VDR_3e-4}
\end{figure}

Based on (\ref{eq:oup_halflife}), for a $VarianceDecayRate=3\times10^{-5}$ and $T_S=150$ time-steps per full episode, the half-life of exploration decay is about 150 episodes as seen in Fig.\ref{fig:OUP_ep350}. However there is an exploration \emph{explosion} after about 650 episodes (Fig.\ref{fig:OUP_ep1300}). As an experiment a severely reduced $VarianceDecayRate=3\times10^{-4}$ was used implying a half-life of just about 15 episodes, however Fig.\ref{fig:OUP_VDR_3e-4} shows no decay even after 1200 episodes.

It is \emph{possible} that the mixed results agree with \citep{b:PLAPPERT}; explicit noise settings are \emph{not} necessary for a continuous space to assist exploration. It must be noted that such results can also be possible due to inexplicable interaction effects of multiple hyperparameters.  

\subsubsection{Sensitivity to random seeds}
Intuitively different random seeds should not affect results of a stable process. According to \citep{b:HENDERSON}, environment stochasticity coupled with stochasticity in the learning process have produced misleading inferences even when results were scientifically averaged across multiple trials.

In conclusion, as stated by \citeauthor{b:HENDERSON} (\citeyear{b:HENDERSON}), one of the possible reasons for the difficulties encountered could be the ``intricate interplay'' of hyperparameters of policy-gradient methods (such as DDPG). 
 
\section{Conclusion}
A process of training a model-free reinforcement learning based controller was outlined. 

Hyperparameter tuning requires significant efforts and patience, for building a stable controller. We proposed Graded Learning, the naive form of Curriculum Learning, which is shown to avoid several challenges associated with RL training. An engineer starts at the lowest task complexity level, allowing one to easily identify the best hyperparameters, reward strategy and reward scales to apply. The control task complexity is then incremented in steps. For most industrial control systems Table \ref{table:DDPG_hyperparameters} should be a good starting point.

Experiments were conducted to evaluate RL against conventional PID control. RL strategy’s trajectory tracking appears to be superior to the PID’s while the PID demonstrates better disturbance rejection as compared to the disturbances appearing on the RL controlled signal. This appears to be the prime limitation of the RL controller and it must be noted that these were also evident in earlier studies by \citep{b:BISCHOFF, b:WANG} and \citep{b:SYAFIIE}.

The PID appeared to lag the reference control signal while the RL controller performed better when challenged to track a control profile that it was not trained on and will demonstrate versatility when applied to different control tasks within the same environment, without having to be retrained.

Overall the RL controlled process appears to promise better process quality, while the PID controlled process will cause a significantly lower stress on the valve operation and result in reduced wear-and-tear.\\

\textbf{Enhancements and Future work}:
The RL controller that was designed needs a mechanism to reduce the oscillatory behavior in the presence of high frequency disturbance with strong amplitudes. For noise at the input and output of the controller a low-pass filter may help reduce the high variance.

Further research is necessary to understand ways of defining objective and reward functions to prevent the noisy RL trajectory behaviour. If this succeeds this will be a better solution than applying a filter, which would otherwise slow down the response.

MATLAB 2019b release includes the Proximal Policy Optimization (PPO) algorithm for continuous control that needs evaluation. PPO is a recent development and is considered as being more stable and better than DDPG \citep{b:HENDERSON}.  

The amalgamation of reinforcement learning, optimal-control and control-systems is extremely exciting. It is hoped that this research will motivate further research to help better understand and popularize the use of reinforcement learning for control-systems.

\section*{Acknowledgement}
This paper is a result of the dissertation submitted to the Coventry University, UK. I am grateful for the guidance of my supervisors Dr. Olivier Haas, Associate Professor and Reader in Applied Control Systems at Coventry University and Dr. Prithvi Sekhar Pagala, Research Specialist at KPIT Technologies. Prof. Dr. Acharya K.N.S must be thanked for instilling an interest in Control Systems through this teaching. The expert comments from reviewers helped improve the quality of this work.




\bibliographystyle{elsarticle-harv} 
\bibliography{RL_for_Valve_Control}

\begin{thebibliography}{38}
\expandafter\ifx\csname natexlab\endcsname\relax\def\natexlab#1{#1}\fi
\providecommand{\url}[1]{\texttt{#1}}
\providecommand{\href}[2]{#2}
\providecommand{\path}[1]{#1}
\providecommand{\DOIprefix}{doi:}
\providecommand{\ArXivprefix}{arXiv:}
\providecommand{\URLprefix}{URL: }
\providecommand{\Pubmedprefix}{pmid:}
\providecommand{\doi}[1]{\href{http://dx.doi.org/#1}{\path{#1}}}
\providecommand{\Pubmed}[1]{\href{pmid:#1}{\path{#1}}}
\providecommand{\bibinfo}[2]{#2}
\ifx\xfnm\relax \def\xfnm[#1]{\unskip,\space#1}\fi
\bibitem[{Bischoff et~al.(2013)Bischoff, Nguyen-Tuong, Koller, Markert and
  Knoll}]{b:BISCHOFF}
\bibinfo{author}{Bischoff, B.}, \bibinfo{author}{Nguyen-Tuong, D.},
  \bibinfo{author}{Koller, T.}, \bibinfo{author}{Markert, H.},
  \bibinfo{author}{Knoll, A.}, \bibinfo{year}{2013}.
\newblock \bibinfo{title}{Learning throttle valve control using policy search},
  in: \bibinfo{booktitle}{Joint European Conference on Machine Learning and
  Knowledge Discovery in Databases}, \bibinfo{organization}{Springer}. pp.
  \bibinfo{pages}{49--64}.
\bibitem[{Bousquet and Elisseeff(2002)}]{b:BOUSQUET}
\bibinfo{author}{Bousquet, O.}, \bibinfo{author}{Elisseeff, A.},
  \bibinfo{year}{2002}.
\newblock \bibinfo{title}{Stability and generalization}.
\newblock \bibinfo{journal}{Journal of machine learning research}
  \bibinfo{volume}{2}.
\bibitem[{di~Capaci and Scali(2018)}]{b:CAPACI}
\bibinfo{author}{di~Capaci, R.B.}, \bibinfo{author}{Scali, C.},
  \bibinfo{year}{2018}.
\newblock \bibinfo{title}{An augmented pid control structure to compensate for
  valve stiction}.
\newblock \bibinfo{journal}{IFAC-PapersOnLine} \bibinfo{volume}{51},
  \bibinfo{pages}{799--804}.
\bibitem[{Choudhury et~al.(2004a)Choudhury, Shah and
  Thornhill}]{b:CHOUDHURY_2004_Quantification}
\bibinfo{author}{Choudhury, M.A.A.S.}, \bibinfo{author}{Shah, S.L.},
  \bibinfo{author}{Thornhill, N.F.}, \bibinfo{year}{2004}a.
\newblock \bibinfo{title}{Detection and quantification of control valve
  stiction}.
\newblock \URLprefix
  \url{http://www.sciencedirect.com/science/article/pii/S1474667017319183}.
\bibitem[{Choudhury et~al.(2004b)Choudhury, Thornhill and
  Shah}]{b:CHOUDHURY_2004_Data_Driven}
\bibinfo{author}{Choudhury, M.A.A.S.}, \bibinfo{author}{Thornhill, N.F.},
  \bibinfo{author}{Shah, S.L.}, \bibinfo{year}{2004}b.
\newblock \bibinfo{title}{A data-driven model for valve stiction}.
\newblock \bibinfo{journal}{IFAC Proceedings Volumes} \bibinfo{volume}{37},
  \bibinfo{pages}{245--250}.
\bibitem[{Deisenroth and Rasmussen(2011)}]{b:PILCO}
\bibinfo{author}{Deisenroth, M.}, \bibinfo{author}{Rasmussen, C.E.},
  \bibinfo{year}{2011}.
\newblock \bibinfo{title}{Pilco: A model-based and data-efficient approach to
  policy search}, in: \bibinfo{booktitle}{Proceedings of the 28th International
  Conference on machine learning (ICML-11)}, pp. \bibinfo{pages}{465--472}.
\bibitem[{Desborough and Miller(2002)}]{b:DESBOROUGH}
\bibinfo{author}{Desborough, L.}, \bibinfo{author}{Miller, R.},
  \bibinfo{year}{2002}.
\newblock \bibinfo{title}{Increasing customer value of industrial control
  performance monitoring—honeywell's experience}.
\bibitem[{Duan et~al.(2016)Duan, Chen, Houthooft, Schulman and Abbeel}]{b:DUAN}
\bibinfo{author}{Duan, Y.}, \bibinfo{author}{Chen, X.},
  \bibinfo{author}{Houthooft, R.}, \bibinfo{author}{Schulman, J.},
  \bibinfo{author}{Abbeel, P.}, \bibinfo{year}{2016}.
\newblock \bibinfo{title}{Benchmarking deep reinforcement learning for
  continuous control}.
\bibitem[{Garnier et~al.(2003)Garnier, Mensler and Richard}]{b:GARNIER}
\bibinfo{author}{Garnier, H.}, \bibinfo{author}{Mensler, M.},
  \bibinfo{author}{Richard, A.}, \bibinfo{year}{2003}.
\newblock \bibinfo{title}{Continuous-time model identification from sampled
  data: implementation issues and performance evaluation}.
\newblock \bibinfo{journal}{International Journal of Control}
  \bibinfo{volume}{76}, \bibinfo{pages}{1337--1357}.
\bibitem[{Hardt et~al.(2015)Hardt, Recht and Singer}]{b:HARDT}
\bibinfo{author}{Hardt, M.}, \bibinfo{author}{Recht, B.},
  \bibinfo{author}{Singer, Y.}, \bibinfo{year}{2015}.
\newblock \bibinfo{title}{Train faster, generalize better: Stability of
  stochastic gradient descent}.
\bibitem[{He and Wang(2010)}]{b:HE_2010}
\bibinfo{author}{He, Q.}, \bibinfo{author}{Wang, J.}, \bibinfo{year}{2010}.
\newblock \bibinfo{title}{Valve stiction modeling: First-principles vs
  data-drive approaches}, in: \bibinfo{booktitle}{Proceedings of the 2010
  American Control Conference}, \bibinfo{publisher}{IEEE}. pp.
  \bibinfo{pages}{3777--3782}.
\bibitem[{Henderson et~al.(2017)Henderson, Islam, Bachman, Pineau, Precup and
  Meger}]{b:HENDERSON}
\bibinfo{author}{Henderson, P.}, \bibinfo{author}{Islam, R.},
  \bibinfo{author}{Bachman, P.}, \bibinfo{author}{Pineau, J.},
  \bibinfo{author}{Precup, D.}, \bibinfo{author}{Meger, D.},
  \bibinfo{year}{2017}.
\newblock \bibinfo{title}{Deep reinforcement learning that matters} .
\bibitem[{Howell and Best(2000)}]{b:HOWELL}
\bibinfo{author}{Howell, M.N.}, \bibinfo{author}{Best, M.C.},
  \bibinfo{year}{2000}.
\newblock \bibinfo{title}{On-line pid tuning for engine idle-speed control
  using continuous action reinforcement learning automata}.
\newblock \bibinfo{journal}{Control Engineering Practice} \bibinfo{volume}{8},
  \bibinfo{pages}{147--154}.
\bibitem[{IEEE-GlobalSpec(1998)}]{b:ISA}
\bibinfo{author}{IEEE-GlobalSpec}, \bibinfo{year}{1998}.
\newblock \bibinfo{title}{Control valves}.
\bibitem[{Lewis et~al.(2012)Lewis, Vrabie and Vamvoudakis}]{b:FRANK}
\bibinfo{author}{Lewis, F.}, \bibinfo{author}{Vrabie, D.},
  \bibinfo{author}{Vamvoudakis, K.}, \bibinfo{year}{2012}.
\newblock \bibinfo{title}{Reinforcement learning and feedback control: Using
  natural decision methods to design optimal adaptive controllers}.
\newblock \URLprefix \url{https://ieeexplore.ieee.org/document/6315769}.
\bibitem[{Lillicrap et~al.(2015)Lillicrap, Hunt, Pritzel, Heess, Erez, Tassa,
  Silver and Wierstra}]{b:LCRAP}
\bibinfo{author}{Lillicrap, T.P.}, \bibinfo{author}{Hunt, J.J.},
  \bibinfo{author}{Pritzel, A.}, \bibinfo{author}{Heess, N.},
  \bibinfo{author}{Erez, T.}, \bibinfo{author}{Tassa, Y.},
  \bibinfo{author}{Silver, D.}, \bibinfo{author}{Wierstra, D.},
  \bibinfo{year}{2015}.
\newblock \bibinfo{title}{Continuous control with deep reinforcement learning}.
\newblock \bibinfo{journal}{arXiv preprint arXiv:1509.02971} .
\bibitem[{MathWorks(2019a)}]{b:MATLAB}
\bibinfo{author}{MathWorks}, \bibinfo{year}{2019}a.
\newblock \bibinfo{title}{Mathworks announces release 2019a of matlab and
  simulink}.
\newblock \URLprefix
  \url{https://in.mathworks.com/company/newsroom/mathworks-announces-release-2019a-of-matlab-and-simulink.html}.
\bibitem[{MathWorks(2019b)}]{b:MATLAB_DDPG}
\bibinfo{author}{MathWorks}, \bibinfo{year}{2019}b.
\newblock \bibinfo{title}{Matlab: rlddpgagentoptions}.
\newblock \URLprefix
  \url{https://in.mathworks.com/help/reinforcement-learning/ref/rlddpgagentoptions.html}.
\bibitem[{Murray et~al.(1994)Murray, Li and Sastry}]{b:MURRAY}
\bibinfo{author}{Murray, R.}, \bibinfo{author}{Li, Z.},
  \bibinfo{author}{Sastry, S.S.}, \bibinfo{year}{1994}.
\newblock \bibinfo{title}{PID Control}. \bibinfo{publisher}{CRC Press}.
\newblock A Mathematical Introduction to Robotic Manipulation, pp.
  \bibinfo{pages}{301--322}.
\bibitem[{Narvekar et~al.(2020)Narvekar, Peng, Leonetti, Sinapov, Taylor and
  Stone}]{b:NARVEKAR}
\bibinfo{author}{Narvekar, S.}, \bibinfo{author}{Peng, B.},
  \bibinfo{author}{Leonetti, M.}, \bibinfo{author}{Sinapov, J.},
  \bibinfo{author}{Taylor, M.E.}, \bibinfo{author}{Stone, P.},
  \bibinfo{year}{2020}.
\newblock \bibinfo{title}{Curriculum learning for reinforcement learning
  domains: A framework and survey}.
\newblock \bibinfo{journal}{arXiv preprint arXiv:2003.04960} .
\bibitem[{Olivecrona et~al.(2017)Olivecrona, Blaschke, Engkvist and
  Chen}]{b:DENOVO}
\bibinfo{author}{Olivecrona, M.}, \bibinfo{author}{Blaschke, T.},
  \bibinfo{author}{Engkvist, O.}, \bibinfo{author}{Chen, H.},
  \bibinfo{year}{2017}.
\newblock \bibinfo{title}{Molecular de-novo design through deep reinforcement
  learning}.
\newblock \bibinfo{journal}{Journal of Cheminformatics} \bibinfo{volume}{9}.
\bibitem[{openai.com(2018)}]{b:OPENAI}
\bibinfo{author}{openai.com}, \bibinfo{year}{2018}.
\newblock \bibinfo{title}{Learning dexterity}.
\newblock \URLprefix \url{https://openai.com/blog/learning-dexterity/}.
\bibitem[{Plappert et~al.(2017)Plappert, Houthooft, Dhariwal, Sidor, Chen,
  Chen, Asfour, Abbeel and Andrychowicz}]{b:PLAPPERT}
\bibinfo{author}{Plappert, M.}, \bibinfo{author}{Houthooft, R.},
  \bibinfo{author}{Dhariwal, P.}, \bibinfo{author}{Sidor, S.},
  \bibinfo{author}{Chen, R.Y.}, \bibinfo{author}{Chen, X.},
  \bibinfo{author}{Asfour, T.}, \bibinfo{author}{Abbeel, P.},
  \bibinfo{author}{Andrychowicz, M.}, \bibinfo{year}{2017}.
\newblock \bibinfo{title}{Parameter space noise for exploration}.
\newblock \bibinfo{journal}{arXiv preprint arXiv:1706.01905} .
\bibitem[{Popinchalk(2006)}]{b:POPINCHALK}
\bibinfo{author}{Popinchalk, S.}, \bibinfo{year}{2006}.
\newblock \bibinfo{title}{Building accurate, realistic simulink models}.
\newblock \URLprefix
  \url{https://in.mathworks.com/company/newsletters/articles/building-accurate-realistic-simulink-models.html}.
\bibitem[{Portelas et~al.(2019)Portelas, Colas, Hofmann and
  Oudeyer}]{b:PORTELAS}
\bibinfo{author}{Portelas, R.}, \bibinfo{author}{Colas, C.},
  \bibinfo{author}{Hofmann, K.}, \bibinfo{author}{Oudeyer, P.},
  \bibinfo{year}{2019}.
\newblock \bibinfo{title}{Teacher algorithms for curriculum learning of deep
  {RL} in continuously parameterized environments}.
\newblock \bibinfo{journal}{CoRR} \bibinfo{volume}{abs/1910.07224}.
\newblock \URLprefix \url{http://arxiv.org/abs/1910.07224}.
\bibitem[{Schoknecht and Riedmiller(1999)}]{b:SCHOKNECHT}
\bibinfo{author}{Schoknecht, R.}, \bibinfo{author}{Riedmiller, M.},
  \bibinfo{year}{1999}.
\newblock \bibinfo{title}{Using reinforcement learning for engine control}.
\newblock \bibinfo{journal}{IEE Conference Publication} .
\bibitem[{Shoukat~Choudhury et~al.(2005)Shoukat~Choudhury, Thornhill and
  Shah}]{b:CHOUDHURY_2005_Modelling}
\bibinfo{author}{Shoukat~Choudhury, M.A.A.}, \bibinfo{author}{Thornhill, N.F.},
  \bibinfo{author}{Shah, S.L.}, \bibinfo{year}{2005}.
\newblock \bibinfo{title}{Modelling valve stiction}.
\newblock \URLprefix
  \url{http://www.sciencedirect.com/science/article/pii/S0967066104001145}.
\bibitem[{Song et~al.(2019)Song, Jiang, Tu, Du and Neyshabur}]{b:SONG}
\bibinfo{author}{Song, X.}, \bibinfo{author}{Jiang, Y.}, \bibinfo{author}{Tu,
  S.}, \bibinfo{author}{Du, Y.}, \bibinfo{author}{Neyshabur, B.},
  \bibinfo{year}{2019}.
\newblock \bibinfo{title}{Observational overfitting in reinforcement learning}
  \URLprefix \url{https://arxiv.org/abs/1912.02975}.
\bibitem[{Southampton.ac.uk()}]{b:TRAIN}
\bibinfo{author}{Southampton.ac.uk}, .
\newblock \bibinfo{title}{Ground vibration and ground-borne noise from trains}.
\newblock \URLprefix
  \url{https://www.southampton.ac.uk/engineering/research/groups/dynamics/rail/ground_vibration.page}.
\bibitem[{Sutton and Barto(2018)}]{b:BARTO}
\bibinfo{author}{Sutton, R.}, \bibinfo{author}{Barto, A.},
  \bibinfo{year}{2018}.
\newblock \bibinfo{title}{Reinforcement Learning: An Introduction}.
\newblock \bibinfo{edition}{2nd. edition} ed., \bibinfo{publisher}{The MIT
  Press}, \bibinfo{address}{Cambridge, England}.
\bibitem[{Syafiie et~al.(2008)Syafiie, Vilas, Garcia, Tadeo, Alonso and
  Martinez}]{b:SYAFIIE}
\bibinfo{author}{Syafiie, S.}, \bibinfo{author}{Vilas, C.},
  \bibinfo{author}{Garcia, M.R.}, \bibinfo{author}{Tadeo, F.},
  \bibinfo{author}{Alonso, A.A.}, \bibinfo{author}{Martinez, E.},
  \bibinfo{year}{2008}.
\newblock \bibinfo{title}{Intelligent control based on reinforcement learning
  for batch thermal sterilization of canned foods}.
\newblock \URLprefix
  \url{http://www.sciencedirect.com/science/article/pii/S1474667016395027}.
\bibitem[{Tedrake(2009)}]{b:TEDRAKE}
\bibinfo{author}{Tedrake, R.}, \bibinfo{year}{2009}.
\newblock \bibinfo{title}{Analytical Optimal Control with the
  Hamilton-Jacobi-Bellman Sufficiency Theorem}.
  \bibinfo{publisher}{Massachusetts Institute of Technology}.
\newblock Underactuated Robotics: Learning, Planning, and Control for Efficient
  and Agile Machines Course Notes for MIT 6.832, pp. \bibinfo{pages}{74--82}.
\bibitem[{Uhlenbeck and Ornstein(1930)}]{b:OUP}
\bibinfo{author}{Uhlenbeck, G.E.}, \bibinfo{author}{Ornstein, L.S.},
  \bibinfo{year}{1930}.
\newblock \bibinfo{title}{On the theory of the brownian motion}.
\newblock \bibinfo{journal}{Physical review} \bibinfo{volume}{36},
  \bibinfo{pages}{823}.
\bibitem[{Vitelli and Nayebi(2016)}]{b:CARMA}
\bibinfo{author}{Vitelli, M.}, \bibinfo{author}{Nayebi, A.},
  \bibinfo{year}{2016}.
\newblock \bibinfo{title}{Carma : A deep reinforcement learning approach to
  autonomous driving} .
\bibitem[{Wang et~al.(2018)Wang, Velswamy and Huang}]{b:WANG}
\bibinfo{author}{Wang, Y.}, \bibinfo{author}{Velswamy, K.},
  \bibinfo{author}{Huang, B.}, \bibinfo{year}{2018}.
\newblock \bibinfo{title}{A novel approach to feedback control with deep
  reinforcement learning}.
\newblock \URLprefix
  \url{http://www.sciencedirect.com/science/article/pii/S2405896318319177}.
\bibitem[{Weiss et~al.(2016)Weiss, Khoshgoftaar and Wang}]{b:KARL}
\bibinfo{author}{Weiss, K.}, \bibinfo{author}{Khoshgoftaar, T.M.},
  \bibinfo{author}{Wang, D.}, \bibinfo{year}{2016}.
\newblock \bibinfo{title}{A survey of transfer learning}.
\newblock \bibinfo{journal}{Journal of Big data} \bibinfo{volume}{3},
  \bibinfo{pages}{9}.
\bibitem[{Weng(2020)}]{b:WENG_CURRICULUM}
\bibinfo{author}{Weng, L.}, \bibinfo{year}{2020}.
\newblock \bibinfo{title}{Curriculum for reinforcement learning}.
\newblock \bibinfo{journal}{lilianweng.github.io/lil-log} \URLprefix
  \url{https://lilianweng.github.io/lil-log/2020/01/29/curriculum-for-reinforcement-learning.html}.
\bibitem[{Zhang et~al.(2018)Zhang, Vinyals, Munos and Bengio}]{b:ZHANG}
\bibinfo{author}{Zhang, C.}, \bibinfo{author}{Vinyals, O.},
  \bibinfo{author}{Munos, R.}, \bibinfo{author}{Bengio, S.},
  \bibinfo{year}{2018}.
\newblock \bibinfo{title}{A study on overfitting in deep reinforcement
  learning}.

\end{thebibliography}

\end{document}